\documentclass{article}
\usepackage{iclr2027_conference,times}

\usepackage{amsmath,amsfonts,bm}

\def\eqref#1{equation~\ref{#1}}

\def\1{\bm{1}}

\DeclareMathAlphabet{\mathsfit}{\encodingdefault}{\sfdefault}{m}{sl}
\SetMathAlphabet{\mathsfit}{bold}{\encodingdefault}{\sfdefault}{bx}{n}

\usepackage{hyperref}
\usepackage{url}
\usepackage{graphicx}
\usepackage{booktabs}
\usepackage{tabularx}

\title{\texorpdfstring{\raisebox{-0.2em}{\includegraphics[height=1.15em]{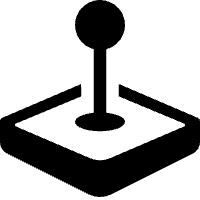}}\hspace{0.25em}}{}ActionRoPE: Action as Coordinates for \\Spatially Consistent Control in \\Planar-Camera Game World Models}

\author{Zhiyang Deng$^{1,2,\dagger}$, Yuzhe Wu$^{1,2,\dagger}$, Zijun Lin$^{1,3,\dagger}$, Yeying Jin$^{1,2,\ddagger*}$ \\
$^{1}$Tencent, $^{2}$National University of Singapore, $^{3}$Nanyang Technological University}

\iclrfinalcopy

\newcommand{\ourmethod}{ActionRoPE}

\begin{document}

\maketitle

\lhead{Preprint}
\begingroup
\renewcommand{\thefootnote}{\fnsymbol{footnote}}
\footnotetext[1]{Corresponding author.}
\footnotetext[2]{This work was completed during research internships at Tencent.}
\footnotetext[3]{Project Lead.}
\endgroup

\begin{figure}[htbp]
\centering
\includegraphics[width=\textwidth]{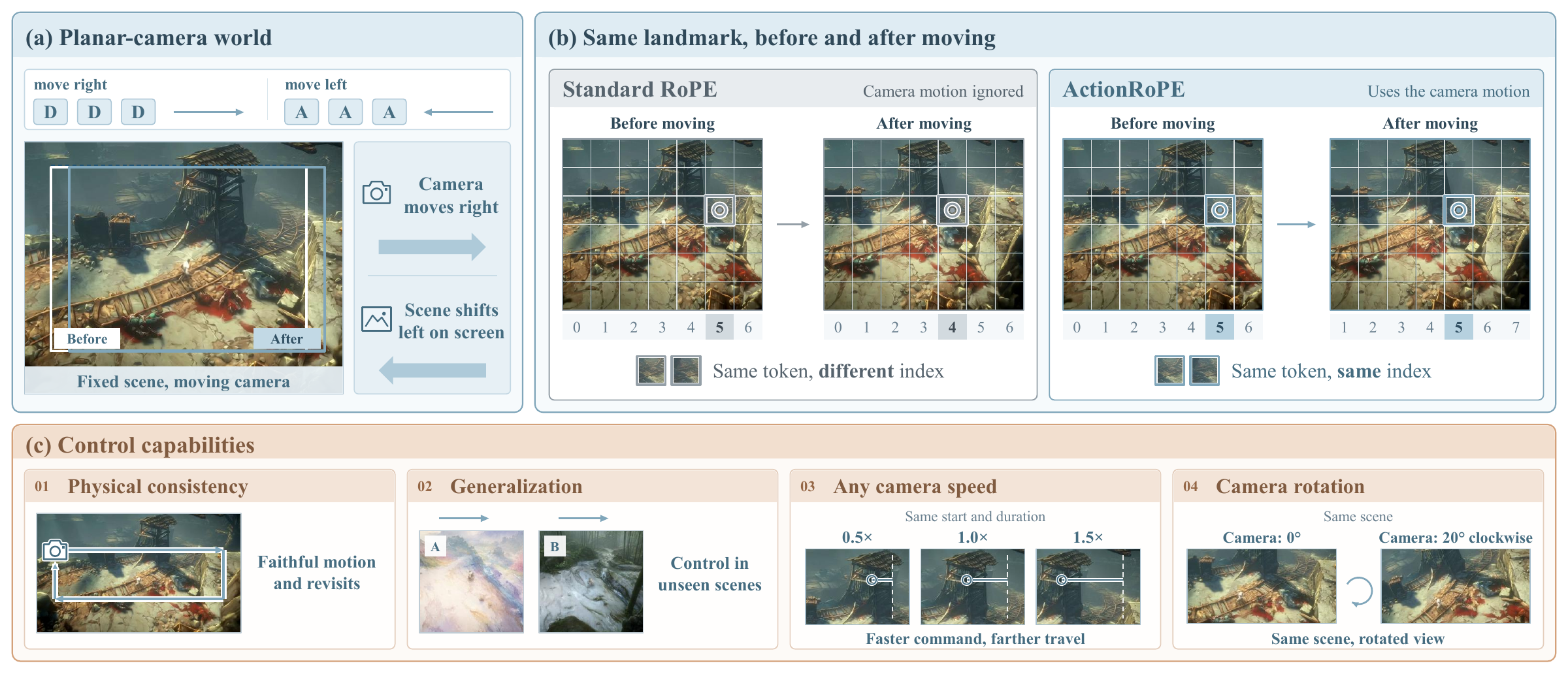}
\vspace{-6pt}
\caption{\textbf{Action as world coordinates.}
(a) In a planar-camera world, each action corresponds to a known translation of the scene.
(b) Standard RoPE re-addresses the same scene token after the camera moves, while \ourmethod{} adds the camera displacement to each token's position, keeping its index across frames.
(c) \ourmethod{} enables precise motion and revisits, control in unseen scenes, and variable camera speed, without adding parameters to the control path.}
\label{fig:teaser}
\end{figure}

\begin{abstract}
Game world models have made rapid progress toward real-time interactive generation. Yet action control is still predominantly implemented as feature conditioning, leaving the model to learn from training data how each action should move the visual world. This formulation is poorly matched to a broad class of \emph{planar-camera worlds}, such as top-down and isometric games. In these worlds, player motion induces a geometrically well-defined planar camera displacement. We observe that the key to spatially consistent control in such worlds is to represent the effect of each action directly in world coordinates. To this end, we introduce \textbf{ActionRoPE}, a parameter-free action interface that converts action-driven camera displacement into world-coordinate rotary positions. As a result, tokens corresponding to the same world location retain the same positional address across time. These world coordinates further expose a spatial asymmetry in generation: regions covered by the initial observation are geometrically known, while newly revealed regions remain to be synthesized. We therefore propose \textbf{Coordinate-Conditioned Noise Scheduling (CNS)}, which assigns token-wise denoising schedules according to this known--new structure. It allows observed regions to stabilize earlier, while new content is generated with reference to them. Compared with four representative action-conditioning mechanisms, our approach follows the commanded motion more faithfully and reduces the revisit drift by $66\%$ over the strongest baseline. The same interface transfers across video backbones from 1.3B to 33B parameters and to out-of-distribution scenes. Project Page: \href{https://teawhite.cn/ActionRoPE/}{\textcolor{blue}{https://teawhite.cn/ActionRoPE/}}
\end{abstract}

\section{Introduction}

Game world models are emerging as a promising paradigm for interactive visual simulation~\citep{bruce2024genie,parkerholder2024genie}. By conditioning video generation on user actions, they allow users to explore and manipulate generated worlds. Recent advances have substantially improved visual fidelity, temporal coherence, and generation efficiency, bringing these models closer to real-time playable environments~\citep{valevski2024gamengen,decart2024oasis,parkerholder2025genie}. Yet realistic video alone is not enough for interaction. The generated world must also respond to user actions accurately and in a spatially consistent way.

Existing game world models typically incorporate user actions as additional conditioning signals. Depending on the architecture, actions may be represented as discrete~\citep{bruce2024genie} or textual~\citep{incantation} tokens, or projected into the model as continuous features~\citep{reactivegwm}. They may also modulate intermediate activations through adaptive normalization~\citep{alayaworld} or enter through dedicated cross-attention modules~\citep{matrixgame3}. Despite their different implementations, these approaches share the same underlying formulation: an action is encoded as a feature, and the generative model is expected to learn how that feature should control the visual world. In other words, both the direction and magnitude of the resulting motion are learned from data rather than explicitly specified by the control interface. This raises a fundamental question: \textit{How should action control be represented when its visual effect is governed by explicit geometry instead of being learned from conditioning features?}

This question becomes particularly relevant in \emph{planar-camera worlds}, such as top-down and isometric games. In these worlds, the player often stays near the screen center while the camera follows the player, causing the static scene to translate almost rigidly in the image plane. In this setting, the effect of an action is largely geometric. Once the action-to-camera scale is calibrated, the action directly determines both the direction and magnitude of the camera displacement. Yet existing interfaces still require the model to learn this geometric translation from data. As a result, the generated motion tends to regress toward motion patterns observed during training, failing to precisely follow the requested displacement. More fundamentally, feature-based action representations do not provide the model with an explicit world coordinate system. Without such coordinates, the model cannot directly tell whether a current token refers to a world location that has already been observed. When the camera returns to a previously seen location, the scene there may no longer look the same. The problem is therefore not merely insufficient conditioning capacity, but a mismatch between feature-based action representations and the geometric structure of planar-camera worlds.

To address this mismatch, we propose \textbf{ActionRoPE}, a parameter-free action interface for planar-camera world models. Instead of encoding an action as an additional feature, ActionRoPE writes the resulting camera displacement directly into the positional coordinates of video tokens. Concretely, it converts screen-space RoPE~\citep{su2024roformer} into action-aligned world-space RoPE, so that tokens corresponding to the same world location retain the same positional address across time (Figure~\ref{fig:teaser}). In this way, the action directly specifies the direction and magnitude of the resulting motion, enabling precise and geometric control. At the same time, the world coordinates naturally reveal which tokens belong to the scene already determined by the first frame and which belong to newly exposed regions. Building on this structure, we further introduce \textbf{Coordinate-Conditioned Noise Scheduling (CNS)}. Unlike applying an undifferentiated denoising schedule to all spatial tokens, CNS assigns token-wise schedules~\citep{chen2024diffusionforcing,ruhe2024rolling,kim2024fifo} according to a known--new partition. Specifically, as the camera moves, tokens inside the first-frame region are denoised and stabilized earlier, while newly revealed tokens keep the original schedule. This allows newly revealed regions to be generated with reference to the already settled scene. Besides a $0.015$M-parameter input channel that feeds the known--new partition to the network, both components add no learnable parameters and leave the underlying attention architecture unchanged.

We evaluate ActionRoPE against four representative action-conditioning strategies on the same backbone and under the same training setting. Across motion scaling, directional compliance, and closed-loop revisit, our method yields more precise and spatially consistent control, reducing the revisit drift by $66\%$ over the strongest baseline. CNS further improves world consistency by reducing the revisit error. The same interface also works on backbones of different sizes~\citep{wan2025,chen2025skyreelsv2,hacohen2024ltx} and transfers to out-of-distribution scenes. This suggests that the coordinate-based control interface is not tied to a particular architecture or training visual domain. Together, our results show that representing action through world coordinates provides a simple yet effective alternative to conventional feature-based control.

To sum up, our contributions are as follows:

\begin{itemize}
\item We introduce \textbf{ActionRoPE}, a parameter-free action-control formulation for planar-camera world models that represents camera motion directly in world-coordinate rotary positions. It turns action control from learned feature conditioning into explicit coordinate alignment.
\item Building on the coordinates exposed by ActionRoPE, we propose \textbf{Coordinate-Conditioned Noise Scheduling (CNS)}, which distinguishes geometrically known and newly revealed regions. It assigns token-wise denoising schedules according to this partition, with only a $0.015$M-parameter input channel and no extra inference passes.
\item We establish an evaluation protocol for action control covering motion fidelity, directional compliance, and closed-loop revisit. On the same backbone, our method outperforms four representative action-conditioning mechanisms and reduces the revisit drift by $66\%$ over the strongest baseline. The same interface also transfers to backbones ranging from \textbf{1.3B} to \textbf{33B} parameters and to out-of-distribution scenes.
\end{itemize}

\section{Related Work}
\label{sec:related}

\subsection{Interactive Game World Models}
\label{sec:related_gwm}

Game world models have evolved from passive gameplay generation toward interactive visual simulation, where future observations are generated in response to user actions~\citep{yu2025position,liu2026interactive}. Genie~\citep{bruce2024genie} learns latent actions from unlabeled gameplay, while subsequent diffusion-based models adopt explicit controls~\citep{alonso2024diffusion}, including keyboard inputs~\citep{valevski2024gamengen,decart2024oasis}, cross-scene control~\citep{yu2025gamefactory}, and real-time streaming generation~\citep{zhang2025matrix,skywork2025matrixgame2,matrixgame3}. Recent work further incorporates richer states and interaction signals~\citep{lin2026stateplay,deng2026worldmind,tong2026scope,chen2026h3world}. Despite these differences, actions are typically injected as conditioning features, such as latent tokens, additive embeddings, language, adaptive normalization, or cross-attention~\citep{bruce2024genie,reactivegwm,incantation,alayaworld,matrixgame3}. The resulting visual motion is therefore still learned from data.

Our work studies a setting where the action-to-motion relation is explicitly geometric. In planar-camera worlds, calibrated player actions determine planar camera displacement, allowing us to represent control directly through world coordinates rather than learned action features.

\subsection{Geometry-Aware Positional Encoding}
\label{sec:related_pe}

Rotary positional embeddings (RoPE) are widely used to encode spatio-temporal token coordinates in video DiTs~\citep{su2024roformer,wan2025}. Recent work has further incorporated camera geometry directly into positional representations rather than treating it only as an external condition~\citep{miyato2023gta,li2025cameras}. ReDirector~\citep{redirector}, ReRoPE~\citep{rerope}, SCoPE~\citep{scope}, and Matrix-Game~3.5~\citep{matrixgame35} encode camera transformations, viewing rays, or projection geometry into RoPE-like coordinates, improving geometric consistency under viewpoint changes. These methods, however, are designed for viewpoint control in general 3D scenes and take the camera pose or trajectory as input.

Our method adopts the same geometric view of positional encoding but applies it specifically to action control. Instead of encoding an external 3D camera pose, ActionRoPE directly converts the action-driven planar displacement into world coordinates, so that the same world location retains the same positional address throughout the camera motion.

\subsection{Adaptive Diffusion Scheduling}
\label{sec:related_sched}

Standard diffusion models apply a shared noise level across all tokens at each denoising step. Several works instead assign different noise levels to different tokens. Diffusion Forcing~\citep{chen2024diffusionforcing}, Rolling Diffusion~\citep{ruhe2024rolling}, FIFO-Diffusion~\citep{kim2024fifo}, and streaming generators~\citep{chen2025skyreelsv2,sandai2025magi} vary noise levels across temporal positions, and recent image-generation methods further explore spatially adaptive schedules~\citep{papa2026asyncpatch,schusterbauer2026denoising}. Related inpainting approaches exploit known regions through explicit resampling or content injection~\citep{lugmayr2022repaint,wan2025,hacohen2024ltx}. These methods show that different regions can benefit from different denoising schedules, but the distinction is typically defined by time, an external mask, or explicit observed content.

In contrast, our method uses the world coordinates provided by ActionRoPE to define this distinction directly. Locations that are already covered by the initial observation are treated as geometrically known, while newly revealed locations still remain uncertain. CNS then uses this known--new structure to assign token-wise denoising schedules, allowing known regions to stabilize earlier without explicitly copying or reinserting any observed content.

\section{Method}
\label{sec:method}

\begin{figure}[t]
\centering
\includegraphics[width=\textwidth]{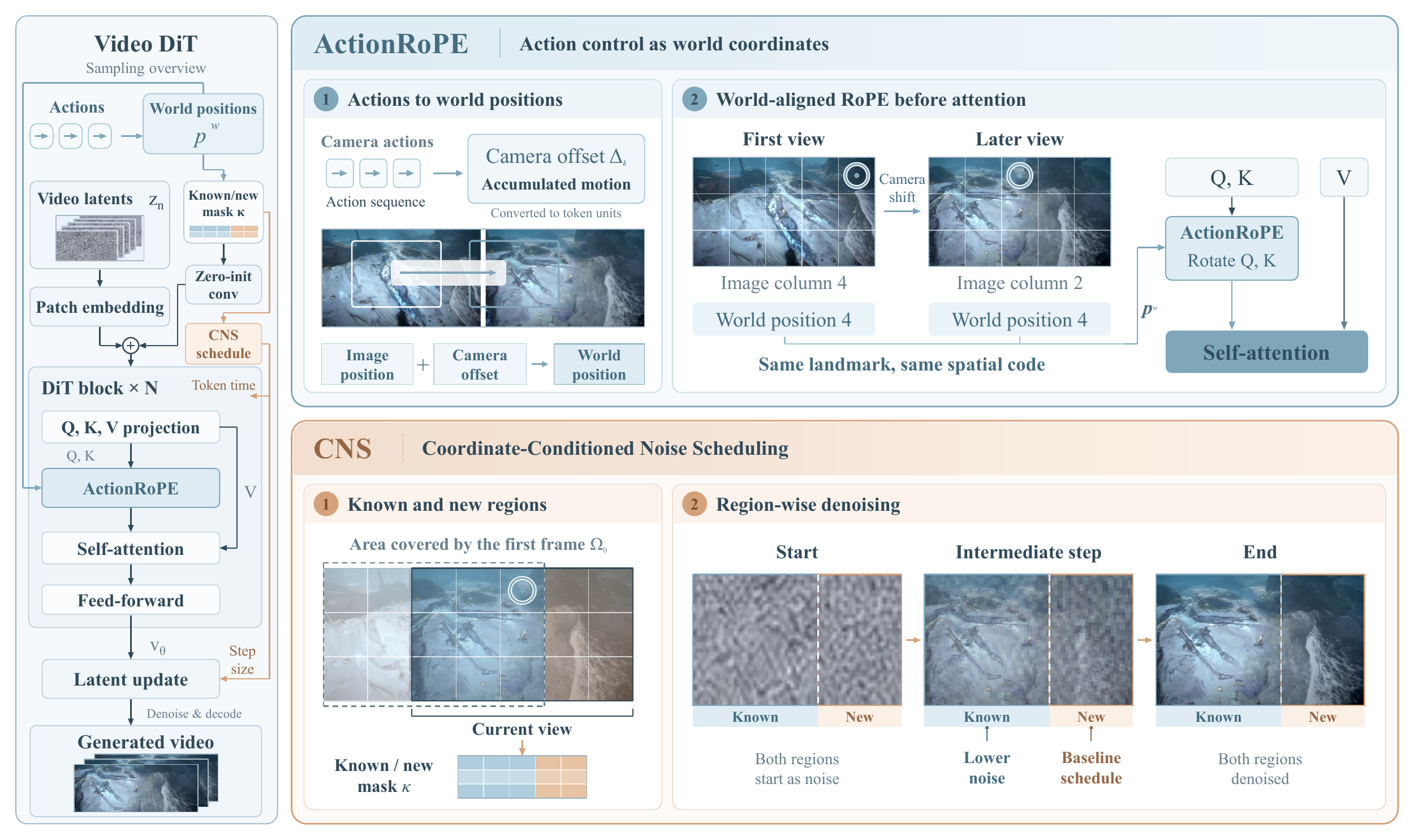}
\caption{\textbf{Overview of \ourmethod{}.}
The left panel shows the video DiT, where \ourmethod{} replaces the screen-relative RoPE before self-attention and the known/new mask $\kappa$ is added to the patch embedding through a zero-initialized convolution. By binding camera motion to world coordinates, \ourmethod{} keeps the same positional index for the same scene location across views, as illustrated in the upper-right panel. The same coordinates reveal which regions are known or newly revealed, and CNS assigns them different denoising schedules, as shown in the lower-right panel.}
\label{fig:method}
\end{figure}

As shown in Figure~\ref{fig:method}, we propose \ourmethod{} to turn a pretrained image-to-video diffusion transformer into an action-controlled world model without adding a control pathway.
We first formalize planar-camera worlds, in which the effect of an action is a camera displacement (Section~\ref{sec:pcw}).
We then write this displacement into the rotary positions, so that a token is addressed by where it is in the world rather than where it is on the screen (Section~\ref{sec:arope}).
The same coordinates also separate known regions from newly revealed ones.
Building on this partition, we introduce Coordinate-Conditioned Noise Scheduling (CNS), which denoises known tokens earlier than newly revealed ones during both training and sampling (Section~\ref{sec:cns}).
Finally, Section~\ref{sec:train-infer} summarizes training and inference.

\subsection{Planar-Camera Worlds}
\label{sec:pcw}

We consider a class of interactive game worlds in which the dominant camera motion is approximately planar, including top-down and isometric games. Given an initial frame and an action sequence, each action moves the player. The camera then follows the player and the static world translates approximately rigidly in the image plane. Once the action-to-camera scale is calibrated, the visual effect of an action is therefore largely determined by geometry.

Existing action-conditioning mechanisms, however, generally treat actions as additional feature signals, whether as tokens, added features, modulated activations, or cross-attention inputs. Despite their different implementations, these approaches share the same formulation:
\begin{equation}
    a_k
    \rightarrow
    \mathrm{Cond}(a_k)
    \rightarrow
    \mathrm{Video \ Generator}
    \rightarrow
    \Delta_k ,
\end{equation}
where the model learns from data how an action should translate into the resulting visual motion. Consequently, both the direction and the magnitude of the motion are determined implicitly by the training distribution rather than explicitly by the control interface.

This formulation is unnecessarily indirect for planar-camera worlds. Here,
$\Delta_k=(\Delta x_k,\Delta y_k)$ denotes the cumulative camera displacement at temporal cell $k$, i.e., the $k$-th latent frame of the video model. For a static world point, its screen-space location changes with $\Delta_k$, but its location in the underlying world remains unchanged. Thus, rather than learning the effect of an action as a feature transformation, we seek to represent action-induced motion directly through the positional coordinates used by the video model. Our main observation is simple: \emph{in planar-camera worlds, action control can be formulated as coordinate alignment}. This observation leads to ActionRoPE, which replaces screen-relative spatial positions with action-aligned world positions.

\subsection{Action Control as World Coordinates}
\label{sec:arope}

\paragraph{World-coordinate positional encoding.}
Consider a video DiT operating on temporal cell $k$ and spatial token $(i,j)$. Standard spatiotemporal RoPE assigns each token a position
\begin{equation}
    \mathbf{p}_{kij}
    =
    (k,i,j),
\end{equation}
where $(i,j)$ denotes the spatial token location in the current image plane. Since these spatial coordinates are reset for every individual frame, the same world location in the scene receives different positional addresses over time whenever the camera moves.

We instead define an action-aligned world coordinate for each token by explicitly compensating for the cumulative camera displacement $\Delta_k$ at temporal cell $k$:
\begin{equation}
    \mathbf{p}^{\,w}_{kij}
    =
    \left(
    k,\;
    i + \frac{\Delta y_k}{s},\;
    j + \frac{\Delta x_k}{s}
    \right),
    \label{eq:world_coordinate}
\end{equation}
where $s$ denotes the number of image pixels represented by one spatial token. In our Wan2.2 implementation, $s=32$. The temporal coordinate remains unchanged.

We compute the rotary angles directly from the real-valued coordinates in Eq.~\ref{eq:world_coordinate}. As RoPE naturally supports continuous positions, the camera displacement does not need to be quantized to the token grid. We refer to this parameter-free formulation as \textbf{ActionRoPE}.

The key property of ActionRoPE is that tokens corresponding to the same world location share the same spatial positional address across time. Suppose a static object appears at screen coordinate $j$ in the first frame and shifts to $j-\delta$ after camera motion. Since the accumulated displacement $\Delta x_k/s$ increases by $\delta$, both observations are assigned the same world coordinate, $(j-\delta)+\delta=j$.
Cross-frame attention can therefore associate the two observations through positional alignment instead of inferring their relative displacement purely from visual appearance.

\paragraph{Known and newly revealed regions.}
World coordinates also allow us to determine whether a token corresponds to a region already observed in the first frame. Let $\Omega_0$ denote the world-coordinate footprint covered by the first frame. We define a binary knownness field
\begin{equation}
    \kappa_{kij}
    =
    \mathbb{I}
    \left[
        \mathbf{p}^{\,w}_{kij} \in \Omega_0
    \right].
    \label{eq:knownness}
\end{equation}
Tokens with $\kappa_{kij}=1$ correspond to world locations already visible in the first frame, whereas tokens with $\kappa_{kij}=0$ correspond to newly exposed regions that remain to be synthesized.

This partition is entirely geometric and is determined by the given action trajectory itself. It is therefore computed in the same way during training and inference and does not require any auxiliary predictor.
Besides determining the noise schedule in Section~\ref{sec:cns}, $\kappa$ is also given to the network as an input through a zero-initialized convolution added to the patch embedding.
This adds only $0.015$M parameters, leaves the pretrained model unchanged at initialization, and lets each token directly identify whether it belongs to a known or a newly revealed region.

\subsection{Coordinate-Conditioned Noise Scheduling}
\label{sec:cns}

ActionRoPE aligns previously observed world locations across time, and also exposes an asymmetry that standard diffusion and flow-matching schedules ignore. Tokens inside the first-frame footprint $\Omega_0$ correspond to content that has already been observed, whereas tokens outside it correspond to genuinely unknown content. Standard schedules nevertheless apply the same noise level to every spatial location. We therefore propose \textbf{Coordinate-Conditioned Noise Scheduling (CNS)}, which adapts the denoising trajectory of each token according to its knownness $\kappa$, so that known regions settle first and newly revealed content is generated with reference to them.

\paragraph{Shift-based token-wise schedule.}
We build CNS on the shifted schedule of the underlying flow-matching model, $\sigma(u;S)=Su/\bigl(1+(S-1)u\bigr)$, where $u\in[0,1]$ is linear time and $S$ the global shift. Given a global noise level $\sigma$, we recover its linear time $u=\sigma/\bigl(S-(S-1)\sigma\bigr)$ and assign every token its own shift and noise level,
\begin{equation}
    S_{kij}=S\left(\frac{S_{\mathrm{known}}}{S}\right)^{\kappa_{kij}},
    \qquad
    \sigma_{kij}=\sigma(u;S_{kij})=\frac{S_{kij}\,u}{1+(S_{kij}-1)\,u},
    \label{eq:cns}
\end{equation}
with $S_{\mathrm{known}}<S$. For newly revealed regions ($\kappa=0$) we have $S_{kij}=S$ and CNS reduces exactly to the original schedule; for known regions ($\kappa=1$) we have $S_{kij}=S_{\mathrm{known}}$ and the token enters low-noise regimes earlier. All tokens share the endpoints $\sigma_{kij}(0)=0$ and $\sigma_{kij}(1)=1$, so they start from the same noisy initialization and converge to the same clean endpoint. Known regions simply stabilize earlier, and newly revealed regions are generated with reference to them.

\paragraph{Training with spatially varying noise.}
For a sampled global noise level $\sigma$, we noise each spatial location $(k,i,j)$ with its own $\sigma_{kij}$, keeping the standard rectified-flow target and making the timestep embedding token-dependent:
\begin{equation}
    \mathbf{x}_{\sigma}=\left(1-\sigma_{kij}\right)\mathbf{x}_0+\sigma_{kij}\,\boldsymbol{\epsilon},
    \qquad
    \mathbf{v}^{*}=\boldsymbol{\epsilon}-\mathbf{x}_0,
    \qquad
    t_{kij}=1000\,\sigma_{kij}.
    \label{eq:cns_noise}
\end{equation}
Following the standard image-to-video setup, the first temporal cell remains a clean anchor and is excluded from the loss. We additionally up-weight the loss on newly revealed regions, since their content is generated rather than retrieved through coordinate alignment.

\paragraph{Token-wise sampling.}
At inference, CNS requires no additional model evaluations. At step $n$ the model is queried once with the token-wise timesteps, $\mathbf{v}_n=f_{\theta}\bigl(\mathbf{x}_n,\,t^{(n)}_{kij},\,\mathbf{p}^{\,w}_{kij},\,\kappa_{kij}\bigr)$, and each token advances along its own schedule,
\begin{equation}
    \mathbf{x}_{n+1}(k,i,j)=\mathbf{x}_n(k,i,j)+\mathbf{v}_n(k,i,j)\,\bigl(\sigma^{(n+1)}_{kij}-\sigma^{(n)}_{kij}\bigr).
    \label{eq:cns_sampling}
\end{equation}
CNS thus changes only \emph{when} each spatial token is denoised during sampling. It does not warp, copy, or paste any content from the first frame. The content of the known regions is still recovered through the world-coordinate correspondence that is established by ActionRoPE.

\begin{figure}[t]
\centering
\includegraphics[width=\textwidth]{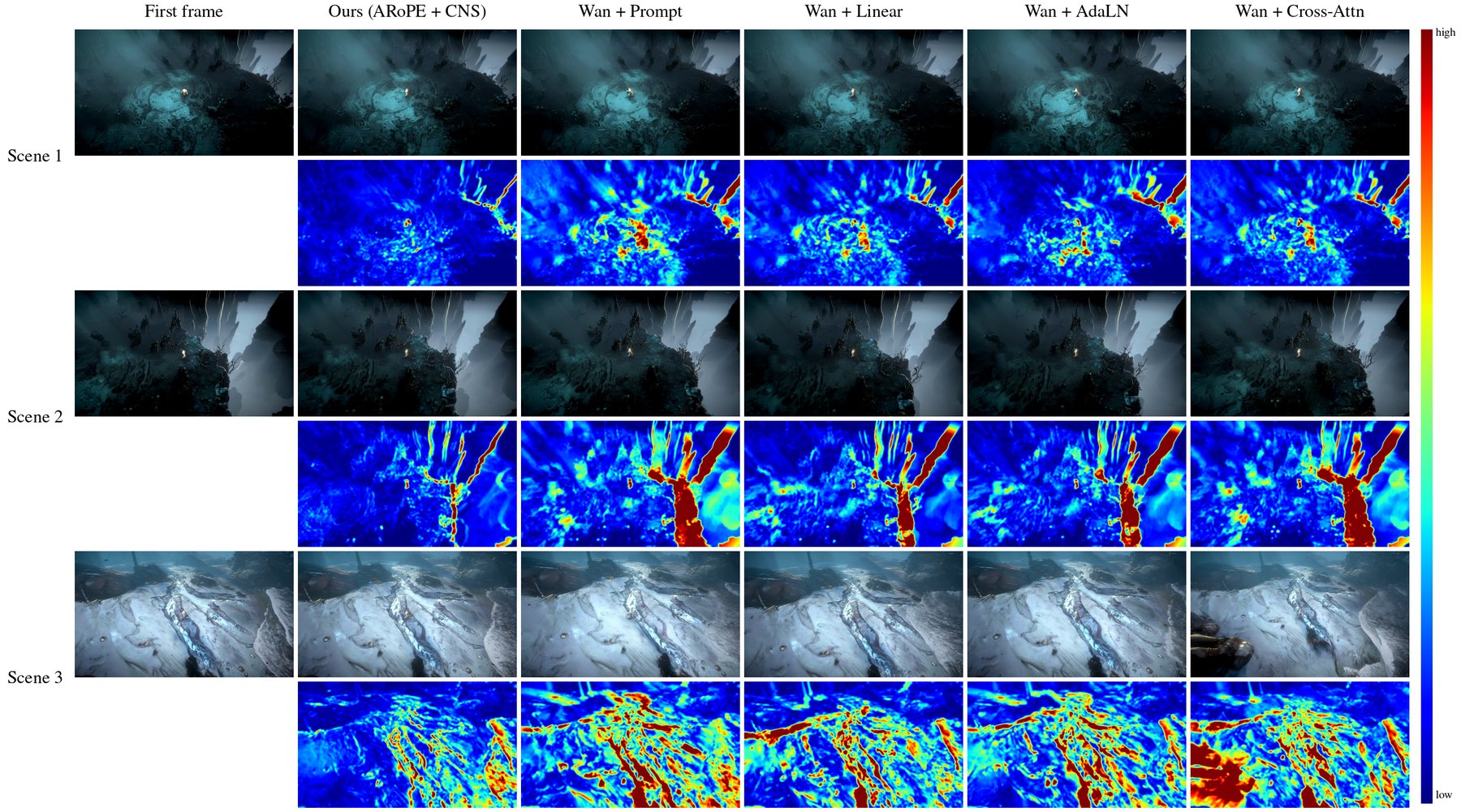}
\caption{\textbf{Qualitative comparison of revisit consistency.} For each scene, the leftmost column shows the first frame. For each method, the top row shows the final frame after a full round trip, and the bottom row shows its absolute error against the first frame (blue indicates low and red indicates high error). Our method keeps the error low across the whole scene, while the four baselines show clearly larger errors that come from shifted or regenerated content.}
\label{fig:qual_revisit}
\end{figure}

\begin{figure}[t]
\centering
\includegraphics[width=\textwidth]{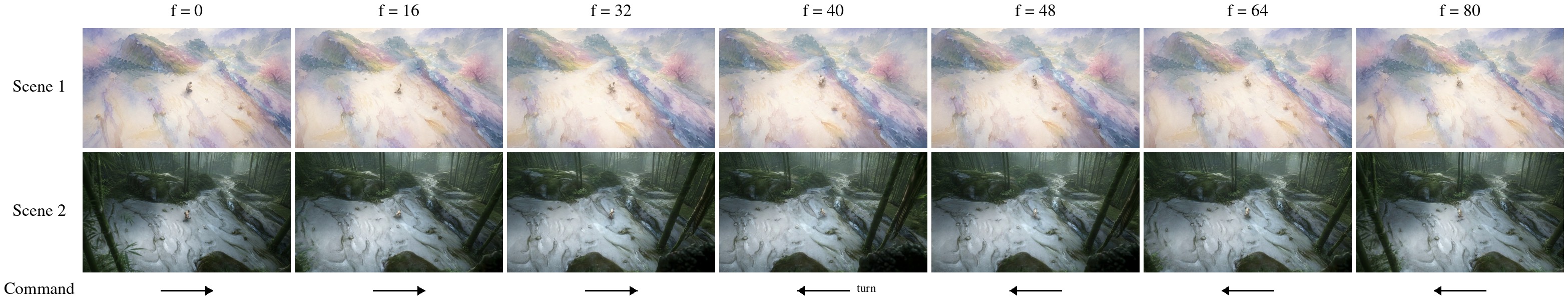}
\caption{\textbf{Control on out-of-distribution first frames.} Each row starts from a first frame in a visual style that is absent from the training game (watercolor and bamboo forest, respectively), and the columns show the generated frames at increasing frame index $f$.}
\label{fig:qual_ood}
\end{figure}

\begin{figure}[t]
\centering
\includegraphics[width=\textwidth]{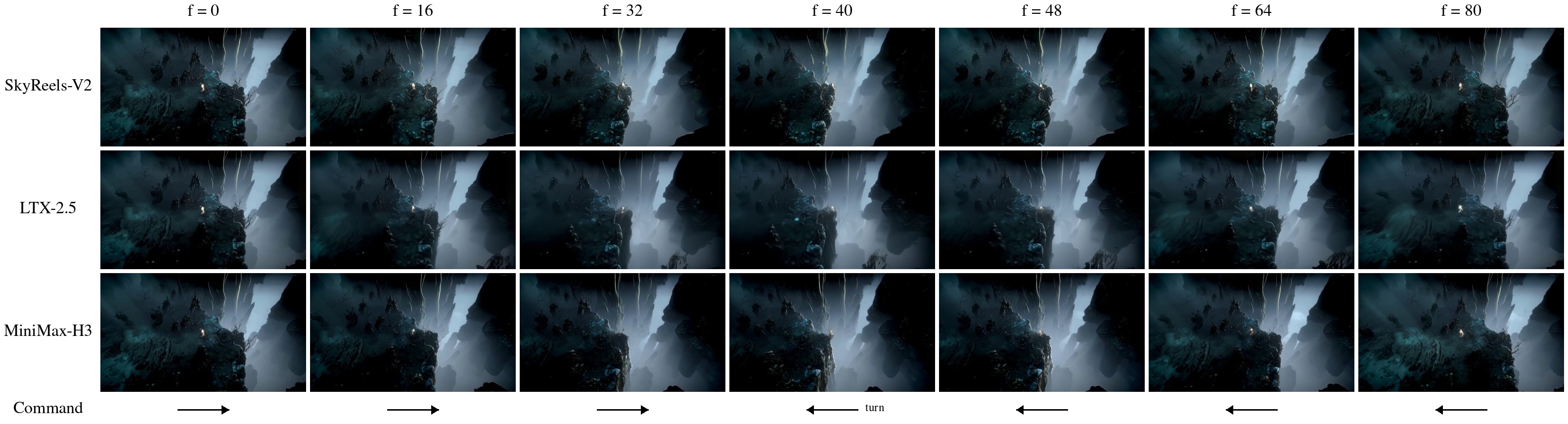}
\caption{\textbf{Control on different backbones.} The same first frame and command are given to \ourmethod{} on SkyReels-V2, LTX-2.5 and MiniMax-H3. The camera moves right and turns back at $f=40$, and the columns show frames at increasing time.}
\label{fig:qual_backbones}
\end{figure}

\subsection{Training and Inference}
\label{sec:train-infer}

During training, we fine-tune the backbone transformer with Eq.~\ref{eq:cns_noise}, taking $\Delta_k$ from the logged player positions and up-weighting the loss on newly revealed tokens by a factor of $2$. Fine-tuning adapts the pretrained model to world-coordinate positions, while the action-to-motion mapping itself is given by geometry. At inference, the user's actions are integrated into $\Delta_k$, which determines the world coordinates and $\kappa$, and the video is sampled with Eq.~\ref{eq:cns_sampling}. For real-time use, we also distill it into a four-step causal student with a key--value cache. Since RoPE attention depends only on the relative positions of tokens, \ourmethod{} carries over to this streaming setting.

\section{Experiments}
\label{sec:exp}

\subsection{Experimental Setup}
\label{sec:exp-setup}

\paragraph{Dataset.}
We build both training and evaluation on \emph{No Rest for the Wicked}, a 2.5D action game rendered from a fixed top-down perspective.
The camera follows the player, so the scene translates almost rigidly in the image plane.
We record $50.9$ hours of automated exploration and cut it into $22{,}782$ clips of $81$ frames at $16$\,fps and $832{\times}480$ resolution.
Of these, $18{,}528$ clips are used for training, and $982$ held-out clips form the evaluation set.
At test time, each model receives only the first frame of an evaluation clip and never its ground-truth continuation.

\begin{table}[t]
\caption{\textbf{Quantitative comparison results with baselines.} On the same Wan backbone, our method achieves the best motion fidelity, directional compliance and revisit consistency. It adds no parameters to the control path and only $0.015$M parameters for the known/new mask channel.}
\vspace{0.5em}
\label{tab:injection}
\centering
\footnotesize
\setlength{\tabcolsep}{2pt}
\begin{tabular*}{\textwidth}{@{\extracolsep{\fill}}lccccccccc@{}}
\toprule
& & \multicolumn{2}{c}{Motion fidelity} & Direction & \multicolumn{5}{c}{Revisit consistency} \\
\cmidrule(lr){3-4}\cmidrule(lr){5-5}\cmidrule(lr){6-10}
Method & Control Params & slope & $R^2$ & angle$\downarrow$ & PSNR$\uparrow$ & SSIM$\uparrow$ & NCC$\uparrow$ & DINO$\uparrow$ & drift$\downarrow$ \\
\midrule
Wan + Prompt     & $0$          & 0.911 & $-$8.46 & 2.19 & 20.98 & 0.613 & 0.684 & 0.852 & 26.6 \\
Wan + Linear     & $+0.18$M     & 0.857 & 0.57    & 3.12 & 20.40 & 0.585 & 0.625 & 0.834 & 35.3 \\
Wan + AdaLN      & $+66.5$M     & 0.784 & $-$0.52 & 1.92 & 21.11 & 0.604 & 0.685 & 0.862 & 22.6 \\
Wan + Cross-Attn & $+98.6$M     & 0.798 & $-$7.18 & 3.05 & 19.48 & 0.529 & 0.515 & 0.817 & 40.2 \\
\midrule
Ours & $0$ & \textbf{0.981} & \textbf{0.952} & \textbf{0.68} & \textbf{23.94} & \textbf{0.691} & \textbf{0.831} & \textbf{0.892} & \textbf{7.8} \\
\bottomrule
\end{tabular*}
\end{table}

\paragraph{Backbones and training.}
We use Wan2.2-TI2V-5B~\citep{wan2025} as our main backbone and implement \ourmethod{}, CNS and all four baselines on it.
All Wan models are fully fine-tuned from the same pretrained weights for $5{,}000$ steps with AdamW, a learning rate of $10^{-5}$ and a batch size of $16$.
To examine whether our interface generalizes across architectures, we further change the backbone to SkyReels-V2-DF 1.3B~\citep{chen2025skyreelsv2}, LTX-2.5 22B~\citep{hacohen2024ltx} and MiniMax-H3 33B~\citep{minimax2026h3} and evaluate under the same setting.

\paragraph{Baselines.}
We compare \ourmethod{} with four representative action-injection mechanisms used by existing game world models, all reimplemented on the same Wan backbone for a fair comparison.
\textbf{Prompt} describes the action in the text prompt of each frame~\citep{incantation}.
\textbf{Linear} projects the action into a feature and adds it to the video tokens in every block~\citep{reactivegwm}.
\textbf{AdaLN} injects the action through adaptive layer normalization~\citep{alayaworld}.
\textbf{Cross-Attn} lets the video tokens attend to action tokens through additional cross-attention layers~\citep{matrixgame3}.
All four baselines receive the same per-frame camera displacement that \ourmethod{} writes into the positional coordinates of the video tokens.

\paragraph{Metrics.}
For each evaluation clip, the model generates $81$ frames from the first frame under a scripted action sequence. We measure the camera motion by matching SIFT features on the background, excluding a $160$-pixel box around the player.
\emph{Motion fidelity} tests whether the camera moves by the commanded amount: on $100$ first frames, we command straight walks at $0.5$--$1.5{\times}$ the nominal speed, and we regress the measured displacement on the commanded one and report the slope and $R^2$, both ideally $1$.
\emph{Directional compliance} tests whether it moves in the commanded direction: on all $982$ first frames, we command a $5$-second straight walk, and we report the median angular error.
\emph{Revisit consistency} tests whether the world stays fixed over time: we command a $2.5$-second walk away followed by a return along the same path back to the start, and we compare the final frame with the first frame by PSNR, SSIM, normalized cross-correlation (NCC), DINOv2 feature similarity~\citep{oquab2024dinov2} and residual geometric drift in pixels.

\subsection{Qualitative Results}
\label{sec:exp-qual}

Figure~\ref{fig:qual_revisit} shows the revisit test on three evaluation scenes. For each method, we show the final frame of the round trip and its error map against the first frame.
With \ourmethod{} and CNS, the final frame closely matches the first frame, and the remaining error concentrates on thin structure edges.
In contrast, the four baselines return with a noticeable offset: their error maps show the scene shifted from its original position, most clearly along the bright path in Scene~2, and Cross-Attn further generates a different layout in the lower-left of Scene~3.

Figure~\ref{fig:qual_ood} shows control on first frames whose styles never appear in the training game.
Given a command that moves right and then turns back at $f=40$, the camera follows the path, and the newly revealed regions are filled in consistently with the style of the first frame.

\subsection{Quantitative Results}
\label{sec:exp-quant}

\paragraph{Comparison with baselines.}
Table~\ref{tab:injection} compares our full model with the four baselines on the same backbone.
Our model outperforms all baselines on every metric without adding any parameters to the control path.
On revisit consistency, it improves over the strongest baseline on each metric by $2.8$\,dB in PSNR, $13\%$ in SSIM and $21\%$ in NCC, and reduces the residual drift by $66\%$.

\begin{table}[t]
\caption{\textbf{Quantitative results across backbones.} With the same action interface, our method controls the camera on backbones ranging from 1.3B to 33B parameters. MiniMax-H3 generates $73$ frames per clip, which are temporally resampled to $81$ frames for evaluation.}
\vspace{0.5em}
\label{tab:backbones}
\centering
\footnotesize
\setlength{\tabcolsep}{3pt}
\begin{tabular*}{\textwidth}{@{\extracolsep{\fill}}lcccccccc@{}}
\toprule
Backbone & slope & $R^2$ & angle$\downarrow$ & PSNR$\uparrow$ & SSIM$\uparrow$ & NCC$\uparrow$ & DINO$\uparrow$ & drift$\downarrow$ \\
\midrule
SkyReels-V2-DF 1.3B & 1.047 & 0.919 & 1.46 & 22.40 & 0.648 & 0.757 & 0.858 & 15.3 \\
Wan2.2-TI2V 5B      & 0.981 & 0.952 & 0.68 & 23.94 & 0.691 & 0.831 & 0.892 & 7.8 \\
LTX-2.5 22B         & 1.014 & 0.845 & 0.59 & 23.55 & 0.717 & 0.810 & 0.800 & 12.8 \\
MiniMax-H3 33B      & 1.082 & 0.861 & 1.90 & 21.97 & 0.647 & 0.726 & 0.831 & 22.0 \\
\bottomrule
\end{tabular*}
\end{table}

\paragraph{Generalization across backbones.}
Table~\ref{tab:backbones} reports the results of our full model on four backbones that differ by more than an order of magnitude in size, and Figure~\ref{fig:qual_backbones} shows the same first frame and command on three of these backbones.
On all four backbones, the measured motion closely follows the commanded speed within $10\%$ and the commanded direction within $2^\circ$, and the median revisit PSNR stays between $22.0$ and $23.9$\,dB.
This indicates that the control comes from the coordinate interface itself rather than from a particular choice of backbone.

\subsection{Ablation Study}
\label{sec:exp-ablation}

\begin{table}[t]
\caption{\textbf{Ablation study.} We remove CNS from our full model on the Wan backbone, while keeping \ourmethod{} in both compared models. Without CNS, all five revisit consistency metrics degrade, and the residual drift increases noticeably from $7.8$ to $11.7$\,px.}
\vspace{0.5em}
\label{tab:ablation}
\centering
\footnotesize

\setlength{\tabcolsep}{16pt}

\begin{tabularx}{\linewidth}{
    @{}
    >{\raggedright\arraybackslash}X
    ccccc
    @{}
}
\toprule
Method & PSNR$\uparrow$ & SSIM$\uparrow$ & NCC$\uparrow$ & DINO$\uparrow$ & drift$\downarrow$ \\
\midrule
Wan + \ourmethod{}
& 22.81 & 0.652 & 0.788 & 0.885 & 11.7 \\

Wan + \ourmethod{} + CNS
& \textbf{23.94} & \textbf{0.691} & \textbf{0.831} & \textbf{0.892} & \textbf{7.8} \\
\bottomrule
\end{tabularx}
\end{table}

We further conduct ablation experiments on each component.
Ablating \ourmethod{} requires replacing it with another action-control mechanism, so the baselines in Table~\ref{tab:injection} can be viewed as ablations of \ourmethod{}.
Here, we therefore focus on the ablation of CNS.
As shown in Table~\ref{tab:ablation}, adding CNS improves all revisit metrics, highlighting the effectiveness of CNS.

\section{Conclusion}
\label{sec:conclusion}

We presented \ourmethod{}, a parameter-free action interface for planar-camera game world models.
It writes action-driven camera motion directly into the rotary positions of video tokens, turning action control from a learned mapping into coordinate alignment.
Building on the world coordinates, CNS lets known regions settle earlier, so that newly revealed content is generated with reference to them.
Experiments show that our method follows the commanded motion more precisely and keeps the world more consistent on revisits than representative action-injection mechanisms.
The same interface also transfers across backbones ranging from 1.3B to 33B parameters.

\subsection*{Acknowledgments}

We would like to express our sincere gratitude to Ruidong Wang and Murphy Zhao for their tremendous support throughout this project.

\bibliography{references}

@article{su2024roformer,
  title   = {{RoFormer:} Enhanced Transformer with Rotary Position Embedding},
  author  = {Su, Jianlin and Ahmed, Murtadha and Lu, Yu and Pan, Shengfeng and Bo, Wen and Liu, Yunfeng},
  journal = {Neurocomputing},
  volume  = {568},
  pages   = {127063},
  year    = {2024}
}

@article{wan2025,
  title   = {Wan: Open and Advanced Large-Scale Video Generative Models},
  author  = {{Team Wan}},
  journal = {arXiv preprint arXiv:2503.20314},
  year    = {2025}
}

@inproceedings{chen2024diffusionforcing,
  title     = {Diffusion Forcing: Next-token Prediction Meets Full-Sequence Diffusion},
  author    = {Chen, Boyuan and Mons{\'o}, Diego Mart{\'i} and Du, Yilun and Simchowitz, Max and Tedrake, Russ and Sitzmann, Vincent},
  booktitle = {Advances in Neural Information Processing Systems (NeurIPS)},
  year      = {2024}
}

@inproceedings{ruhe2024rolling,
  title     = {Rolling Diffusion Models},
  author    = {Ruhe, David and Heek, Jonathan and Salimans, Tim and Hoogeboom, Emiel},
  booktitle = {International Conference on Machine Learning (ICML)},
  year      = {2024}
}

@inproceedings{kim2024fifo,
  title     = {{FIFO-Diffusion}: Generating Infinite Videos from Text without Training},
  author    = {Kim, Jihwan and Kang, Junoh and Choi, Jinyoung and Han, Bohyung},
  booktitle = {Advances in Neural Information Processing Systems (NeurIPS)},
  year      = {2024}
}

@inproceedings{bruce2024genie,
  title     = {Genie: Generative Interactive Environments},
  author    = {Bruce, Jake and Dennis, Michael and Edwards, Ashley and Parker-Holder, Jack and Shi, Yuge and Hughes, Edward and Lai, Matthew and Mavalankar, Aditi and Steigerwald, Richie and Apps, Chris and Aytar, Yusuf and Bechtle, Sarah and Behbahani, Feryal and Chan, Stephanie and Heess, Nicolas and Gonzalez, Lucy and Osindero, Simon and Ozair, Sherjil and Reed, Scott and Zhang, Jingwei and others},
  booktitle = {International Conference on Machine Learning (ICML)},
  year      = {2024}
}

@inproceedings{valevski2024gamengen,
  title     = {Diffusion Models Are Real-Time Game Engines},
  author    = {Valevski, Dani and Leviathan, Yaniv and Arar, Moab and Fruchter, Shlomi},
  booktitle = {International Conference on Learning Representations (ICLR)},
  year      = {2025}
}

@misc{decart2024oasis,
  title  = {Oasis: A Universe in a Transformer},
  author = {{Decart} and {Etched}},
  year   = {2024},
  note   = {\url{https://oasis-model.github.io/}}
}

@article{yu2025gamefactory,
  title   = {{GameFactory:} Creating New Games with Generative Interactive Videos},
  author  = {Yu, Jiwen and Qin, Yiran and Wang, Xintao and Wan, Pengfei and Zhang, Di and Liu, Xihui},
  journal = {arXiv preprint arXiv:2501.08325},
  year    = {2025}
}

@article{skywork2025matrixgame2,
  title   = {{Matrix-Game} 2.0: An Open-Source, Real-Time, and Streaming Interactive World Model},
  author  = {Xianglong He and Chunli Peng and Zexiang Liu and Boyang Wang and Yifan Zhang and Qi Cui and Fei Kang and Biao Jiang and Mengyin An and Yangyang Ren and Baixin Xu and Hao-Xiang Guo and Kaixiong Gong and Size Wu and Wei Li and Xuchen Song and Yang Liu and Yangguang Li and Yahui Zhou},
  journal = {arXiv preprint arXiv:2508.13009},
  year    = {2025}
}

@article{matrixgame3,
  title   = {{Matrix-Game} 3.0: Real-Time and Streaming Interactive World Model with Long-Horizon Memory},
  author  = {Zile Wang and Zexiang Liu and Jiaxing Li and Kaichen Huang and Baixin Xu and Fei Kang and Mengyin An and Peiyu Wang and Biao Jiang and Yichen Wei and Yidan Xietian and Jiangbo Pei and Liang Hu and Boyi Jiang and Hua Xue and Zidong Wang and Haofeng Sun and Wei Li and Wanli Ouyang and Xianglong He and Yang Liu and Yangguang Li and Yahui Zhou},
  journal = {arXiv preprint arXiv:2604.08995},
  year    = {2026}
}

@article{matrixgame35,
  title   = {{Matrix-Game} 3.5: Enhancing Real-Time Streaming Interactive World Models with Patch Memory},
  author  = {Runjia Qian and Zile Wang and Jihai Zhang and Kai Zou and Wei Yu and Jiaxing Li and Zexiang Liu and Yaokun Li and Fei Kang and Kaichen Huang and Mengyin An and Haobo Zhang and Biao Jiang and Jiahua Wang and Haofeng Sun and Yang Liu and Yangguang Li},
  journal = {arXiv preprint arXiv:2608.29910},
  year    = {2026}
}

@article{reactivegwm,
  title   = {{ReactiveGWM:} Steering {NPC} in Reactive Game World Models},
  author  = {Zeqing Wang and Danze Chen and Zhaohu Xing and Zizhao Tong and Yinhan Zhang and Xingyi Yang and Yeying Jin},
  journal = {arXiv preprint arXiv:2605.15256},
  year    = {2026}
}

@article{incantation,
  title   = {{Incantation:} Natural Language as the Action Interface for Multi-Entity Video World Models},
  author  = {Shangwen Zhu and Qianyu Peng and Zhao Pu and Zhilei Shu and Xiangrui Ke and Zhaohu Xing and Zizhao Tong and Zeqing Wang and Xinyu Cui and Huangji Wang and Jian Zhao and Yeying Jin and Fan Cheng and Ruili Feng},
  journal = {arXiv preprint arXiv:2605.18601},
  year    = {2026}
}

@article{alayaworld,
  title   = {{AlayaWorld:} Interactive Long-Horizon World Modeling -- Full Technical Report},
  author  = {{AlayaWorld Team} and Kaipeng Zhang and Chuanhao Li and Yifan Zhan and Yongtao Ge and Yuanyang Yin and Jiaming Tan and Kang He and Liaoyuan Fan and Mingliang Zhai and Ruicong Liu and Xiaojie Xu and Xuangeng Chu and Zhen Li and Zhengyuan Lin and Zhixiang Wang and Zian Meng and Zihui Gao},
  journal = {arXiv preprint arXiv:2607.18367},
  year    = {2026}
}

@article{redirector,
  title   = {{ReDirector:} Creating Any-Length Video Retakes with Rotary Camera Encoding},
  author  = {Byeongjun Park and Byung-Hoon Kim and Hyungjin Chung and Jong Chul Ye},
  journal = {arXiv preprint arXiv:2511.19827},
  year    = {2025}
}

@article{rerope,
  title   = {{ReRoPE:} Repurposing {RoPE} for Relative Camera Control},
  author  = {Chunyang Li and Yuanbo Yang and Jiahao Shao and Hongyu Zhou and Katja Schwarz and Yiyi Liao},
  journal = {arXiv preprint arXiv:2602.08068},
  year    = {2026}
}

@article{scope,
  title   = {{SCoPE:} Sightline-Coordinate Positional Encoding for Video Diffusion Transformers},
  author  = {Minghao Yin and Jiahao Lu and Wenbo Hu and Wang Zhao and Ying Shan and Kai Han},
  journal = {arXiv preprint arXiv:2606.27345},
  year    = {2026}
}

@inproceedings{lugmayr2022repaint,
  title     = {RePaint: Inpainting using Denoising Diffusion Probabilistic Models},
  author    = {Lugmayr, Andreas and Danelljan, Martin and Romero, Andres and Yu, Fisher and Timofte, Radu and Van Gool, Luc},
  booktitle = {IEEE/CVF Conference on Computer Vision and Pattern Recognition (CVPR)},
  year      = {2022}
}

@article{chen2025skyreelsv2,
  title   = {{SkyReels-V2:} Infinite-length Film Generative Model},
  author  = {Chen, Guibin and Lin, Dixuan and Yang, Jiangping and Lin, Chunze and others},
  journal = {arXiv preprint arXiv:2504.13074},
  year    = {2025}
}

@article{hacohen2024ltx,
  title   = {{LTX-Video:} Realtime Video Latent Diffusion},
  author  = {HaCohen, Yoav and Chiprut, Nisan and Brazowski, Benny and Shalem, Daniel and Moshe, Dudu and Richardson, Eitan and Levin, Eran and Shiran, Guy and Zabari, Nir and Gordon, Ori and others},
  journal = {arXiv preprint arXiv:2501.00103},
  year    = {2024}
}

@article{oquab2024dinov2,
  title   = {DINOv2: Learning Robust Visual Features without Supervision},
  author  = {Oquab, Maxime and Darcet, Timoth{\'e}e and Moutakanni, Th{\'e}o and Vo, Huy and Szafraniec, Marc and Khalidov, Vasil and Fernandez, Pierre and Haziza, Daniel and Massa, Francisco and El-Nouby, Alaaeldin and others},
  journal = {Transactions on Machine Learning Research (TMLR)},
  year    = {2024}
}

@article{alonso2024diffusion,
  title   = {Diffusion for World Modeling: Visual Details Matter in {Atari}},
  author  = {Eloi Alonso and Adam Jelley and Vincent Micheli and Anssi Kanervisto and Amos Storkey and Tim Pearce and François Fleuret},
  journal = {arXiv preprint arXiv:2405.12399},
  year    = {2024},
  note    = {NeurIPS 2024}
}

@article{li2025cameras,
  title   = {Cameras as Relative Positional Encoding},
  author  = {Ruilong Li and Brent Yi and Junchen Liu and Hang Gao and Yi Ma and Angjoo Kanazawa},
  journal = {arXiv preprint arXiv:2507.10496},
  year    = {2025},
  note    = {NeurIPS 2025}
}

@article{liu2026interactive,
  title   = {Towards Interactive Video World Modeling: Frontiers, Challenges, Benchmarks, and Future Trends},
  author  = {Jiuming Liu and Chaojun Ni and Mengmeng Liu and Chensheng Peng and Fangjinhua Wang and Sitian Shen and Marc Pollefeys and Masayoshi Tomizuka and Ayush Tewari and Per Ola Kristensson},
  journal = {arXiv preprint arXiv:2606.01164},
  year    = {2026}
}

@article{miyato2023gta,
  title   = {{GTA:} A Geometry-Aware Attention Mechanism for Multi-View Transformers},
  author  = {Takeru Miyato and Bernhard Jaeger and Max Welling and Andreas Geiger},
  journal = {arXiv preprint arXiv:2310.10375},
  year    = {2024},
  note    = {ICLR 2024}
}

@article{papa2026asyncpatch,
  title   = {{AsyncPatch} Diffusion: spatially-flexible image generation},
  author  = {Samuele Papa and Valentin De Bortoli and Guillaume Couairon and Daniel Sýkora and Romuald Elie and Klaus Greff},
  journal = {arXiv preprint arXiv:2606.07079},
  year    = {2026}
}

@misc{parkerholder2024genie,
  title  = {Genie 2: A large-scale foundation world model},
  author = {Jack Parker-Holder and Philip Ball and Jake Bruce and Vibhavari Dasagi and Kristian Holsheimer and Christos Kaplanis and Alexandre Moufarek and Guy Scully and Jeremy Shar and Jimmy Shi and Stephen Spencer and Jessica Yung and Michael Dennis and Sultan Kenjeyev and Shangbang Long and Vlad Mnih and Harris Chan and Maxime Gazeau and Bonnie Li and Fabio Pardo and Luyu Wang and Lei Zhang and Frederic Besse and Tim Harley and Anna Mitenkova and Jane Wang and Jeff Clune and Demis Hassabis and Raia Hadsell and Adrian Bolton and Satinder Singh and Tim Rocktäschel},
  year   = {2024},
  note   = {\url{https://deepmind.google/discover/blog/genie-2-a-large-scale-foundation-world-model/}}
}

@misc{parkerholder2025genie,
  title  = {Genie 3: A new frontier for world models},
  author = {Jack Parker-Holder and Shlomi Fruchter},
  year   = {2025},
  note   = {\url{https://deepmind.google/discover/blog/genie-3-a-new-frontier-for-world-models/}}
}

@article{sandai2025magi,
  title   = {{MAGI-1:} Autoregressive Video Generation at Scale},
  author  = {{Sand.ai} and Hansi Teng and Hongyu Jia and Lei Sun and Lingzhi Li and Maolin Li and Mingqiu Tang and Shuai Han and Tianning Zhang and W. Q. Zhang and Weifeng Luo and Xiaoyang Kang and Yuchen Sun and Yue Cao and Yunpeng Huang and Yutong Lin and Yuxin Fang and Zewei Tao and Zheng Zhang and Zhongshu Wang and Zixun Liu and Dai Shi and Guoli Su and Hanwen Sun and Hong Pan and Jie Wang and Jiexin Sheng and Min Cui and Min Hu and Ming Yan and Shucheng Yin and Siran Zhang and Tingting Liu and Xianping Yin and Xiaoyu Yang and Xin Song and Xuan Hu and Yankai Zhang and Yuqiao Li},
  journal = {arXiv preprint arXiv:2505.13211},
  year    = {2025}
}

@article{schusterbauer2026denoising,
  title   = {Denoising, Fast and Slow: Difficulty-Aware Adaptive Sampling for Image Generation},
  author  = {Johannes Schusterbauer and Ming Gui and Yusong Li and Pingchuan Ma and Felix Krause and Björn Ommer},
  journal = {arXiv preprint arXiv:2604.19141},
  year    = {2026},
  note    = {CVPR 2026}
}

@article{yu2025position,
  title   = {Position: Interactive Generative Video as Next-Generation Game Engine},
  author  = {Jiwen Yu and Yiran Qin and Haoxuan Che and Quande Liu and Xintao Wang and Pengfei Wan and Di Zhang and Xihui Liu},
  journal = {arXiv preprint arXiv:2503.17359},
  year    = {2025}
}

@article{zhang2025matrix,
  title   = {{Matrix-Game:} Interactive World Foundation Model},
  author  = {Yifan Zhang and Chunli Peng and Boyang Wang and Puyi Wang and Qingcheng Zhu and Fei Kang and Biao Jiang and Zedong Gao and Eric Li and Yang Liu and Yahui Zhou},
  journal = {arXiv preprint arXiv:2506.18701},
  year    = {2025}
}

@article{tong2026scope,
  title   = {{SCOPE:} Simulating Cross-game Operations in Playable Environments for {FPS} World Models},
  author  = {Zizhao Tong and Hongfeng Lai and Zeqing Wang and Zhaohu Xing and Kexu Cheng and Haoran Xu and Zhao Pu and Shangwen Zhu and Ruili Feng and Jian Zhao and Yan Zhang and Hao Tang and Yeying Jin and Ling Shao},
  journal = {arXiv preprint arXiv:2605.23345},
  year    = {2026}
}

@article{deng2026worldmind,
  title   = {{WorldMind:} Decoupled Game World Model for State-Aware {NPC} Behavior},
  author  = {Zhiyang Deng and Boran Zhang and Danze Chen and Yeying Jin},
  journal = {arXiv preprint arXiv:2608.21439},
  year    = {2026}
}

@article{lin2026stateplay,
  title   = {{StatePlay:} State-Aware Game World Models for Mechanics-Consistent Generation},
  author  = {Zijun Lin and Zeqing Wang and Cheston Tan and Bihan Wen and Yeying Jin},
  journal = {arXiv preprint arXiv:2607.26754},
  year    = {2026}
}

@article{chen2026h3world,
  title   = {{H3-World:} Turning Language Understanding into World Control},
  author  = {Danze Chen and Zeqing Wang and Ziyue Lin and Xingyi Yang and Yeying Jin},
  journal = {arXiv preprint arXiv:2609.01560},
  year    = {2026}
}

@misc{minimax2026h3,
  title        = {{MiniMax-H3}},
  author       = {{MiniMax}},
  year         = {2026},
  howpublished = {\url{https://huggingface.co/MiniMaxAI/MiniMax-H3}}
}
\bibliographystyle{iclr2027_conference}

\appendix
\clearpage
\newcommand{\apptocsec}[2]{\noindent\hyperref[#1]{\textbf{\makebox[1.6em][l]{\ref*{#1}}#2}}\hfill\textbf{\pageref*{#1}}\par\vspace{0.35em}}
\newcommand{\apptocsub}[2]{\noindent\hspace{1.6em}\hyperref[#1]{\makebox[2.2em][l]{\ref*{#1}}#2}\dotfill\pageref*{#1}\par\vspace{0.15em}}
\begin{center}
{\LARGE\bfseries Supplementary Material for ActionRoPE: Action as Coordinates for Spatially Consistent Control in Planar-Camera Game World Models}
\end{center}
\vspace{1.5em}
\noindent{\large\bfseries Contents}
\vspace{0.8em}

\apptocsec{sec:app-qual}{Additional Qualitative Results}
\apptocsub{sec:app-qual-revisit}{Revisit Comparison}
\apptocsub{sec:app-qual-scaling}{Motion Scaling}
\apptocsub{sec:app-qual-ood}{Out-of-Distribution Scenes}
\apptocsub{sec:app-qual-backbones}{Cross-Backbone Examples}
\apptocsub{sec:app-qual-cns}{Effect of CNS}
\vspace{0.5em}
\apptocsec{sec:app-quant}{Additional Quantitative Results}
\apptocsub{sec:app-quant-ood}{Out-of-Distribution Scenes}
\apptocsub{sec:app-quant-scaling}{Motion Scaling per Speed}
\apptocsub{sec:app-schedule}{Noise Schedule Design}
\apptocsub{sec:app-trainsample}{Training versus Sampling}
\vspace{0.5em}
\apptocsec{sec:app-impl}{Implementation Details}
\apptocsub{sec:app-impl-arope}{ActionRoPE and CNS}
\apptocsub{sec:app-impl-train}{Training and Sampling}
\apptocsub{sec:app-baselines}{Baselines}
\vspace{0.5em}
\apptocsec{sec:app-data}{Dataset}
\vspace{0.5em}
\apptocsec{sec:app-eval}{Evaluation Protocol}
\vspace{0.5em}
\apptocsec{sec:app-limit}{Limitations}

\clearpage
\section{Additional Qualitative Results}
\label{sec:app-qual}
\suppressfloats[t]

\subsection{Revisit Comparison}
\label{sec:app-qual-revisit}

\begin{figure}[t]
\centering
\includegraphics[width=\textwidth]{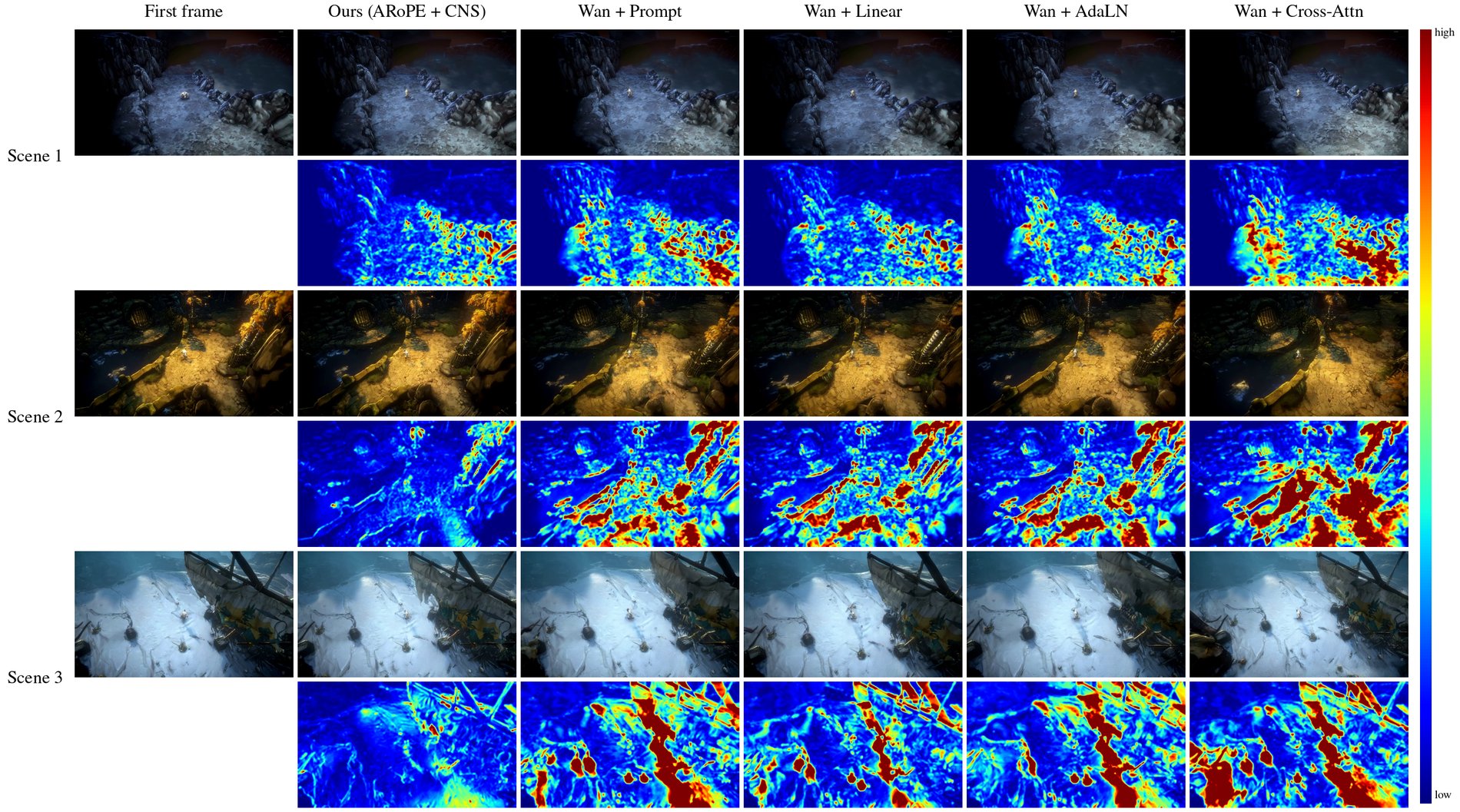}
\caption{\textbf{More revisit comparisons (I).} For each scene, we show the final frame of each method after the round trip and its error map against the first frame.}
\label{fig:app-revisit-1}
\end{figure}

\begin{figure}[t]
\centering
\includegraphics[width=\textwidth]{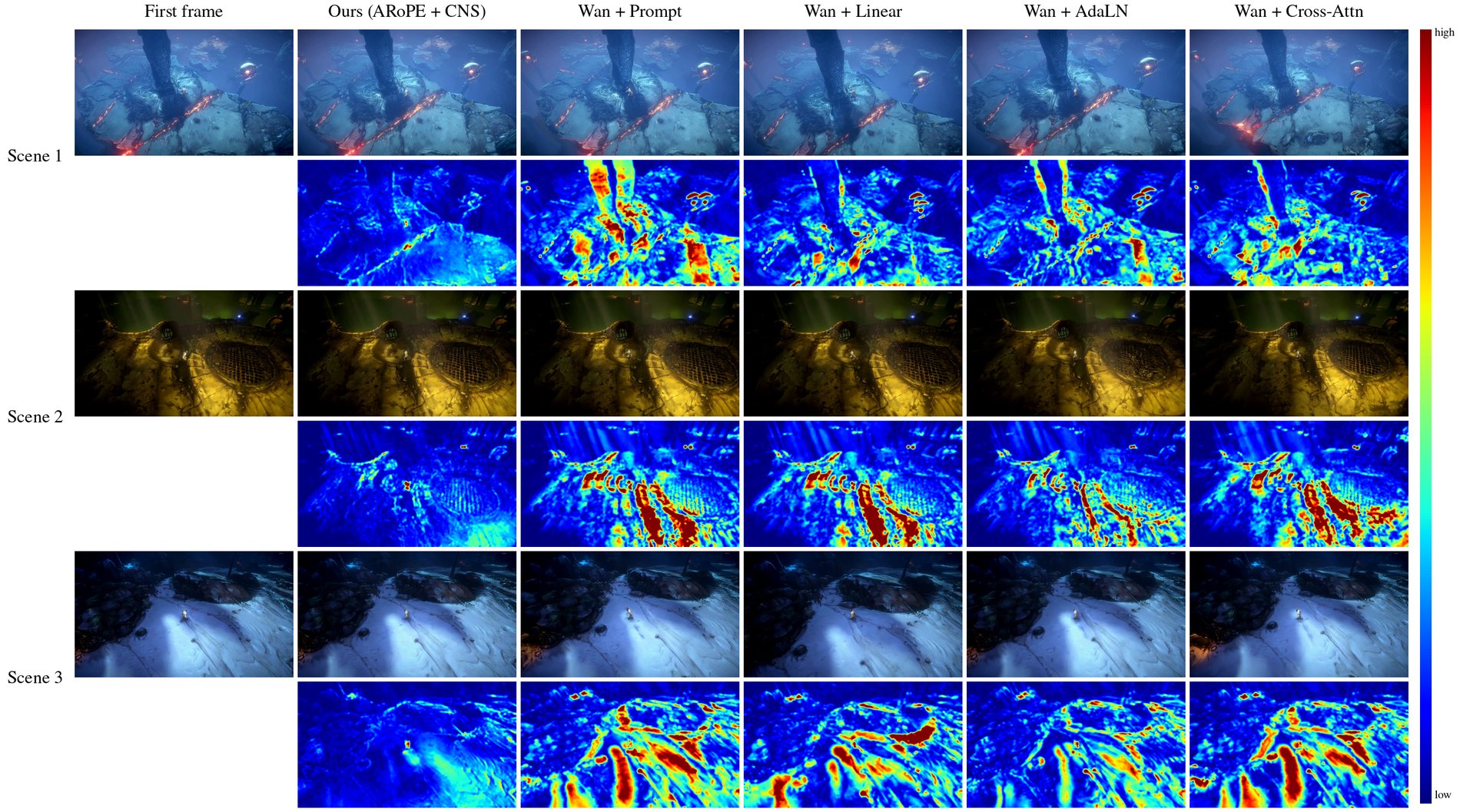}
\caption{\textbf{More revisit comparisons (II).} Same layout as Figure~\ref{fig:app-revisit-1}, on three other evaluation scenes.}
\label{fig:app-revisit-2}
\end{figure}

Figures~\ref{fig:app-revisit-1} and~\ref{fig:app-revisit-2} show the revisit test of Figure~\ref{fig:qual_revisit} on six more evaluation scenes.
All models receive the same command, moving right for $40$ frames and then returning along the same path, and we show the final frame of each method with its error map against the first frame.
With our method, the error stays low over most of the frame and concentrates on thin structure edges.
The four baselines show larger high-error regions where the scene has not returned to its original position, for example on the bright ground in Scene~2 of Figure~\ref{fig:app-revisit-1} and around the rock pillar in Scene~1 of Figure~\ref{fig:app-revisit-2}.

\subsection{Motion Scaling}
\label{sec:app-qual-scaling}

\begin{figure}[t]
\centering
\includegraphics[width=\textwidth]{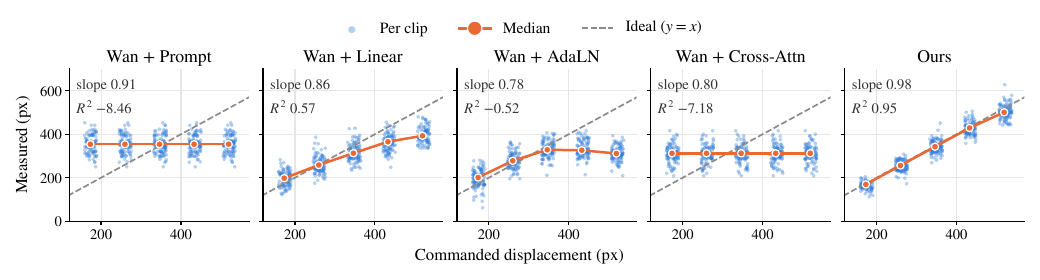}
\caption{\textbf{Measured versus commanded displacement.} Each point is one clip, the orange line connects the medians at each speed, and the dashed line marks exact tracking. The slope and $R^2$ are those of Table~\ref{tab:injection}.}
\label{fig:app-scaling}
\end{figure}

Figure~\ref{fig:app-scaling} plots the measured displacement against the commanded displacement for all $500$ motion-fidelity clips, i.e., $100$ first frames at five speeds.
Our method stays close to the ideal line at every speed.
Linear and AdaLN receive the displacement as input and respond to the commanded speed, but they compress it, and their measured displacement levels off at higher speeds.
Prompt and Cross-Attn receive only the direction of each action (Section~\ref{sec:app-baselines}), so they move by nearly the same amount at every speed, which also explains their negative $R^2$ in Table~\ref{tab:injection}.

\subsection{Out-of-Distribution Scenes}
\label{sec:app-qual-ood}

\begin{figure}[t]
\centering
\includegraphics[width=\textwidth]{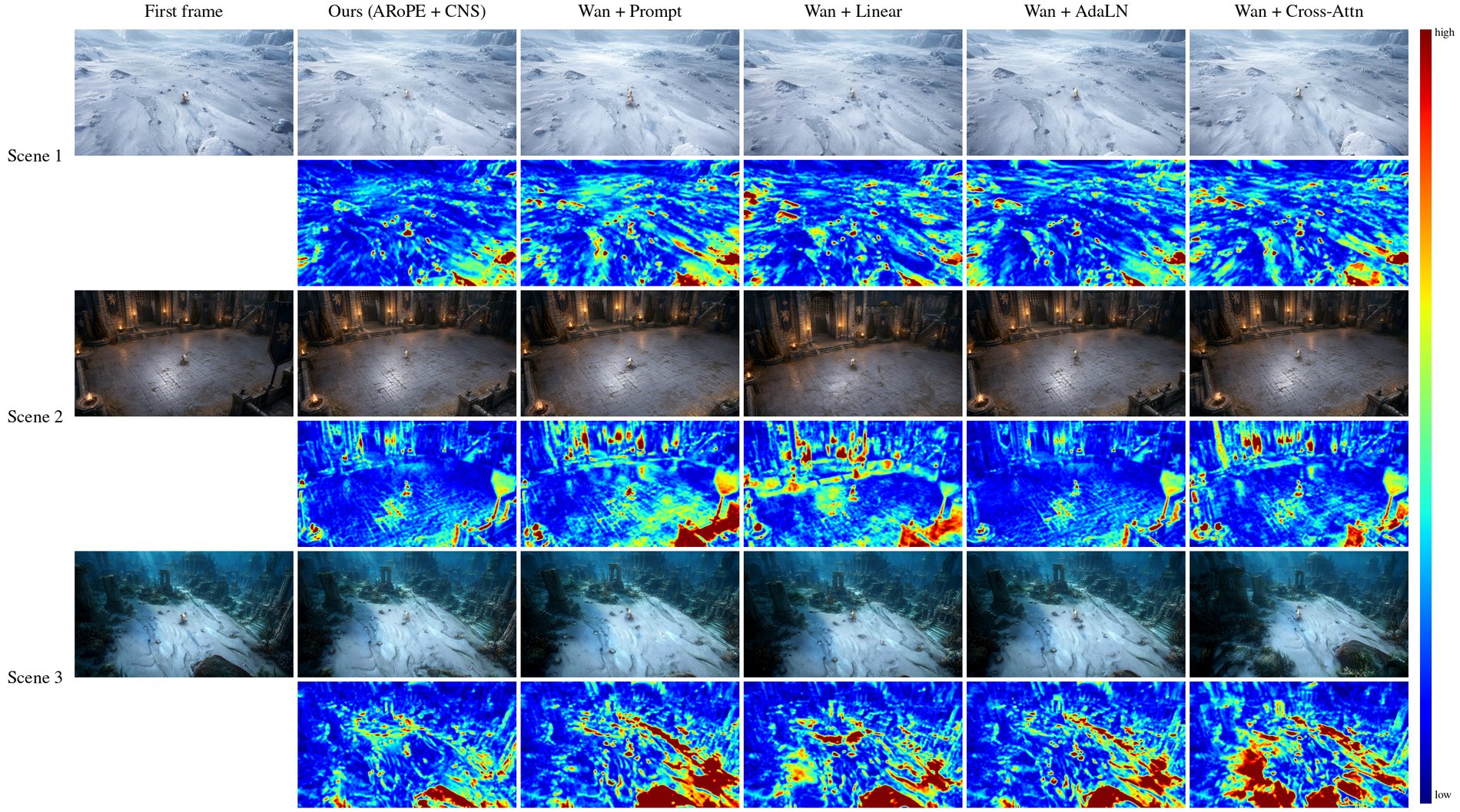}
\caption{\textbf{Revisit comparison on out-of-distribution scenes.} The three first frames show an arctic tundra, a medieval castle and an underwater ruin. Same layout as Figure~\ref{fig:app-revisit-1}.}
\label{fig:app-ood-revisit}
\end{figure}

\begin{figure}[t]
\centering
\includegraphics[width=\textwidth]{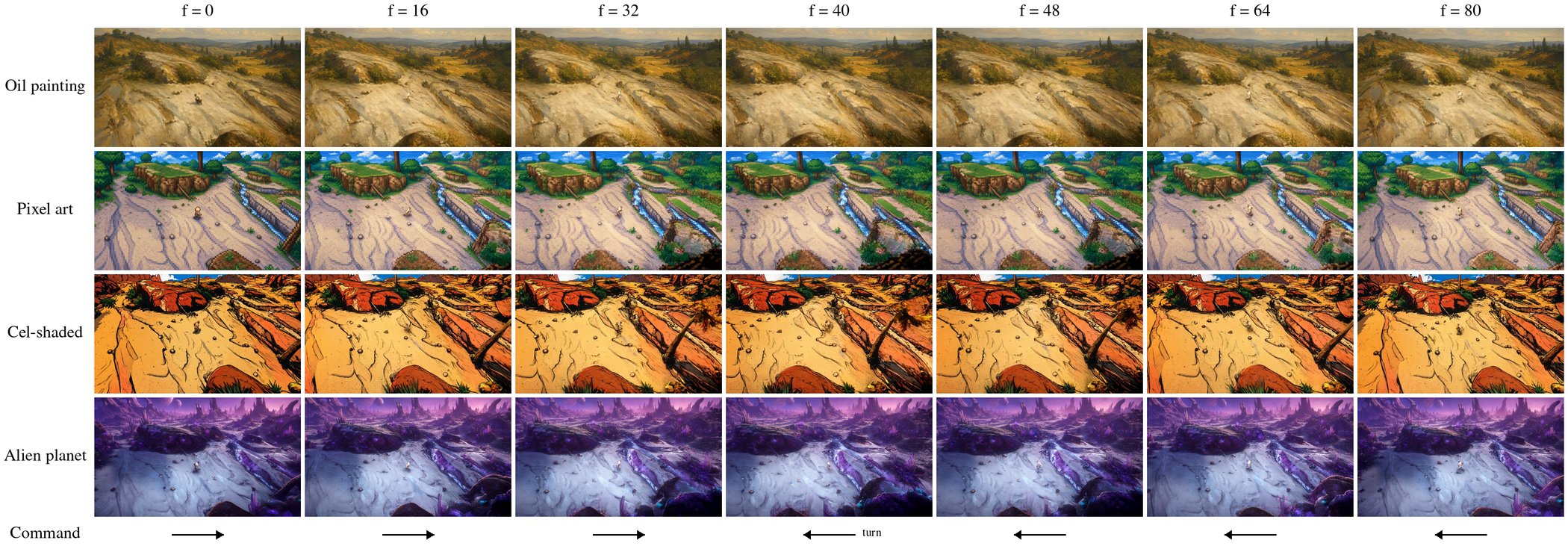}
\caption{\textbf{More out-of-distribution styles.} Our method on four first frames with styles that never appear in the training game, under a command that moves right and turns back at $f{=}40$.}
\label{fig:app-ood-timeline}
\end{figure}

Figure~\ref{fig:app-ood-revisit} repeats the revisit comparison on three first frames whose scenes do not appear in the training game.
As on the evaluation set, the baselines return with larger errors than our method, most visibly on the pillars and in the lower right of Scene~2, while the lower part of Scene~3 changes for all methods, including ours.
Figure~\ref{fig:app-ood-timeline} shows four more styles under the command of Figure~\ref{fig:qual_ood}.
In each case, the camera follows the command and returns close to its starting view at $f{=}80$, and the regions revealed on the right are rendered in the style of the first frame.

\subsection{Cross-Backbone Examples}
\label{sec:app-qual-backbones}

\begin{figure}[t]
\centering
\includegraphics[width=\textwidth]{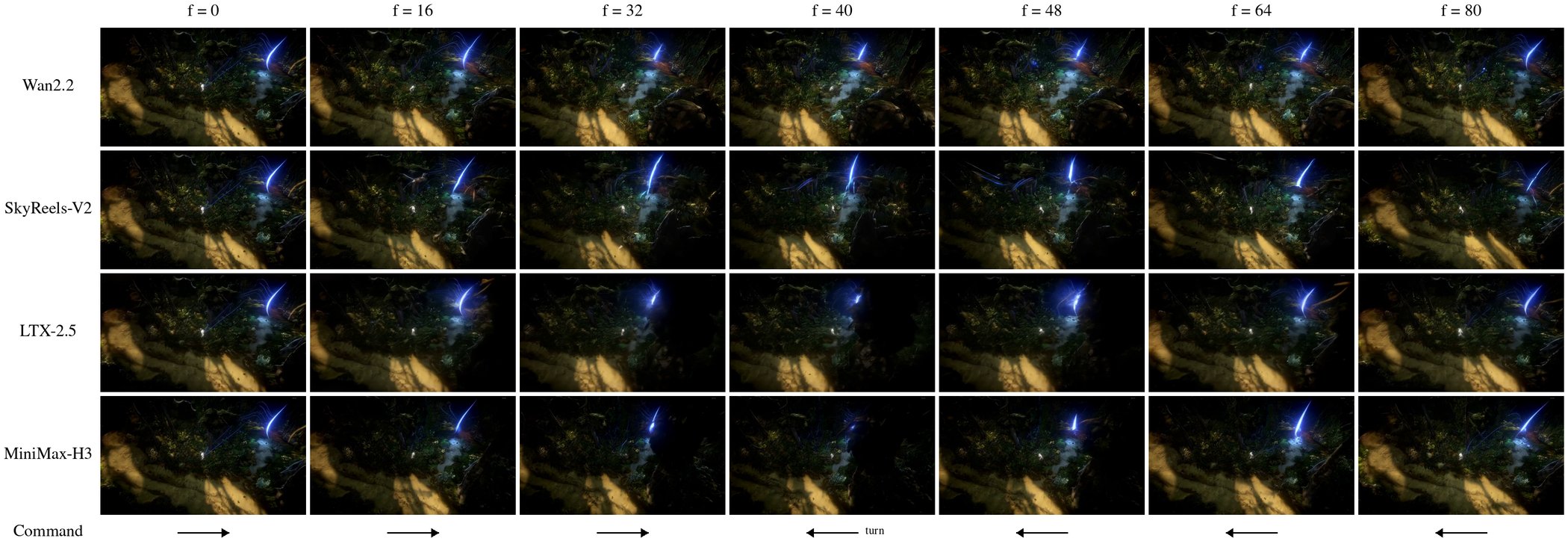}
\caption{\textbf{Cross-backbone example (I).} The same first frame and command on the four backbones.}
\label{fig:app-backbones-1}
\end{figure}

\begin{figure}[t]
\centering
\includegraphics[width=\textwidth]{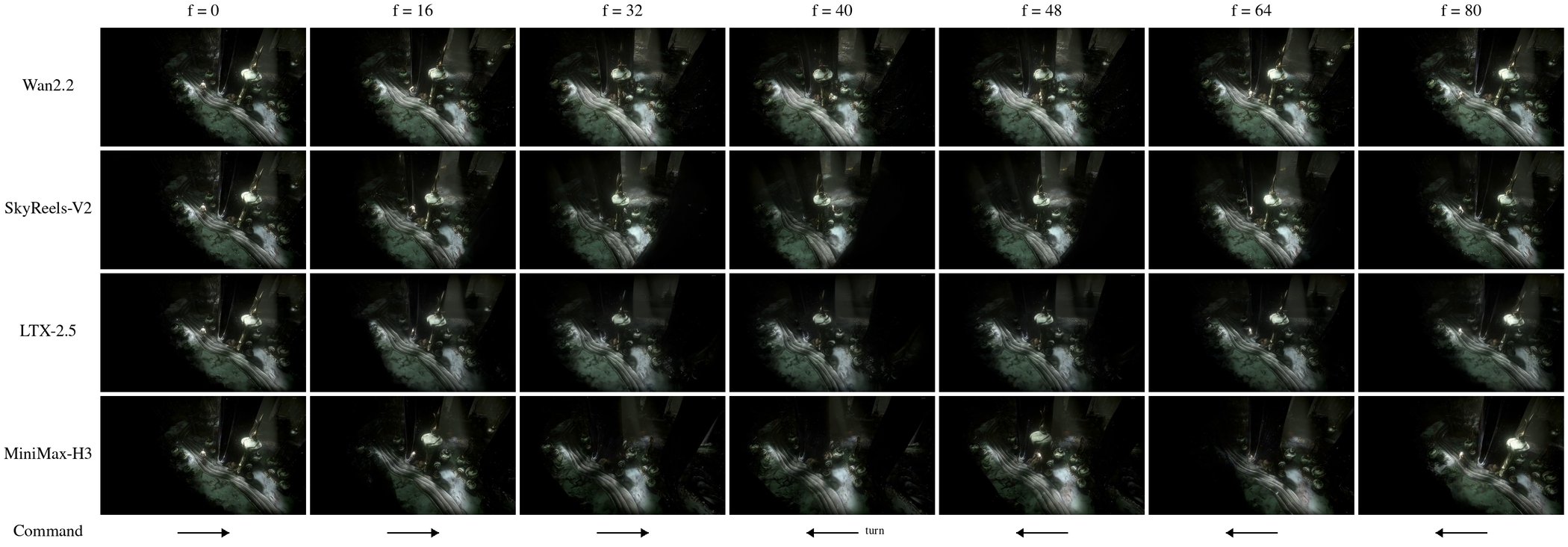}
\caption{\textbf{Cross-backbone example (II).} Same setting as Figure~\ref{fig:app-backbones-1}, on another evaluation scene.}
\label{fig:app-backbones-2}
\end{figure}

Figures~\ref{fig:app-backbones-1} and~\ref{fig:app-backbones-2} apply the same first frame and command to all four backbones, including Wan2.2, which is not shown in Figure~\ref{fig:qual_backbones}.
All four backbones follow the rightward walk, turn back at $f{=}40$ and return close to the starting view at $f{=}80$.
They differ mainly in how they render newly revealed regions, which appear darker for LTX-2.5 and MiniMax-H3 in Figure~\ref{fig:app-backbones-1}.

\subsection{Effect of CNS}
\label{sec:app-qual-cns}

\begin{figure}[t]
\centering
\includegraphics[width=\textwidth]{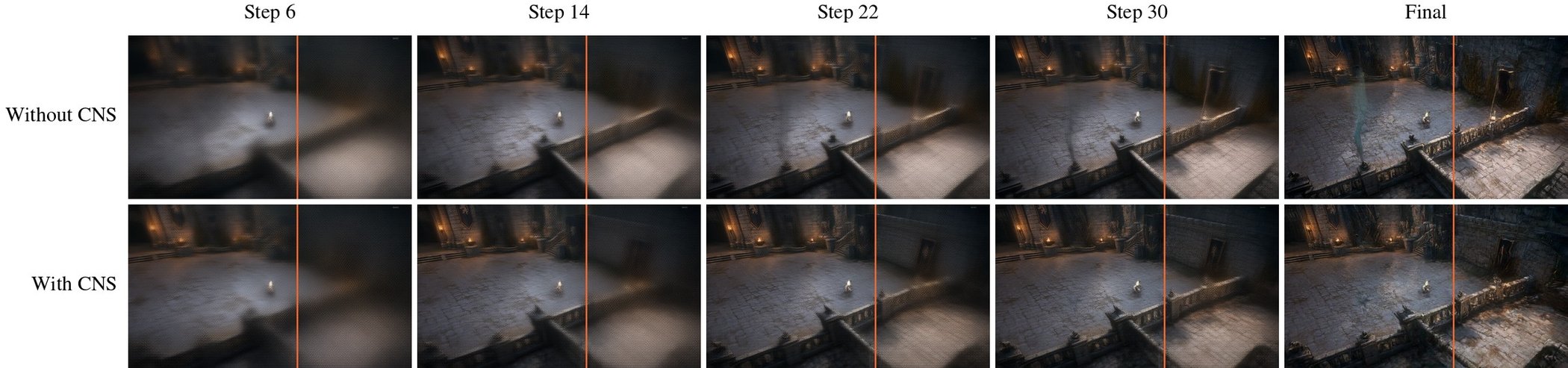}
\caption{\textbf{Sampling with and without CNS.} Clean-frame estimates of the last frame during sampling, under a straight walk to the right. The region to the left of the orange line is covered by the first frame.}
\label{fig:app-cns-progress}
\end{figure}

\begin{figure}[t]
\centering
\includegraphics[width=0.62\textwidth]{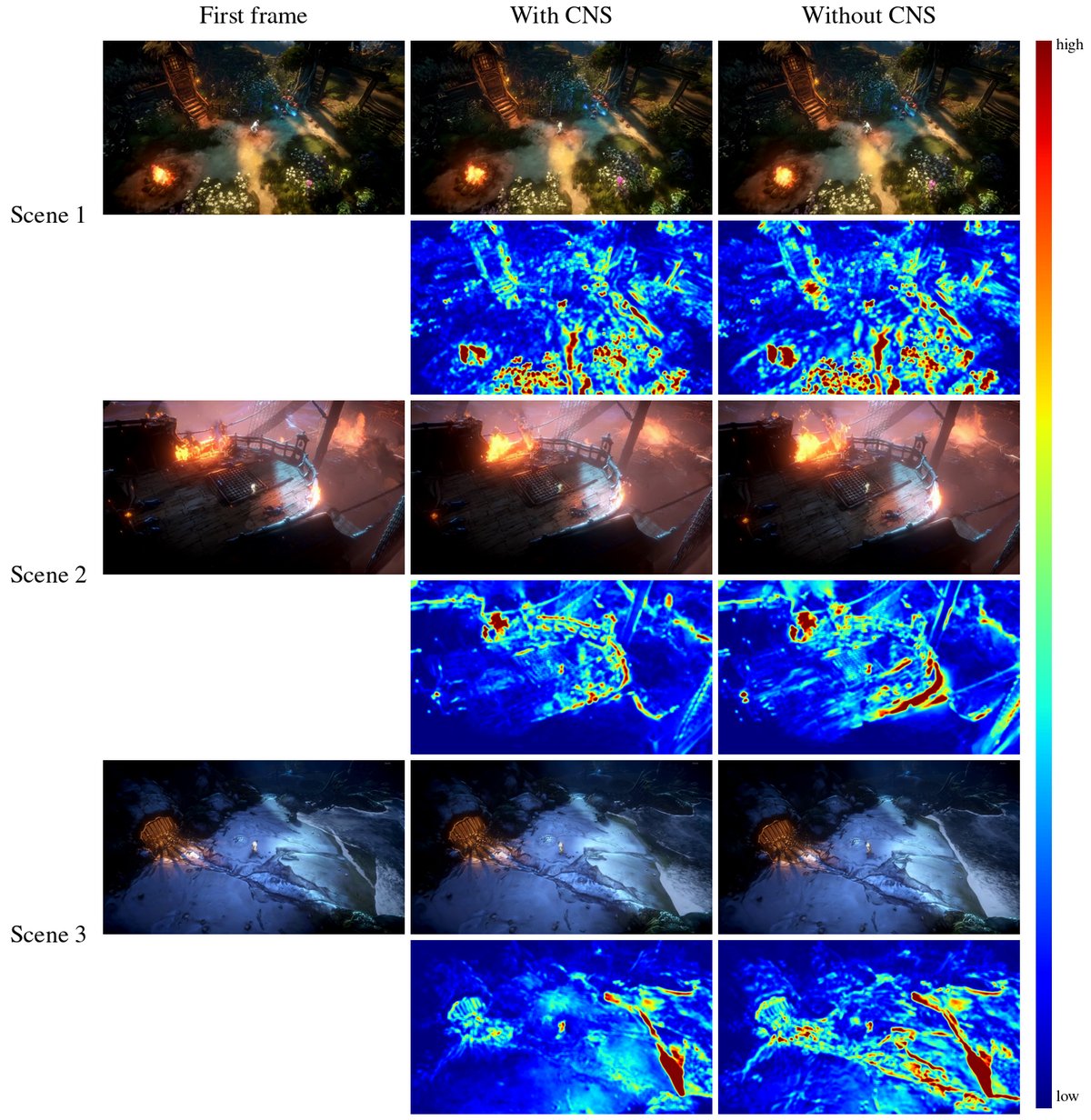}
\caption{\textbf{Revisit with and without CNS.} Final frames after the round trip and their error maps, for our full model and the model trained without CNS.}
\label{fig:app-cns-revisit}
\end{figure}

Figure~\ref{fig:app-cns-progress} visualizes sampling with and without CNS for the same first frame, action and random seed.
At several sampling steps, we decode the current clean-frame estimate of the last frame, where the region to the left of the orange line is covered by the first frame.
In this example, CNS already recovers the floor tiles of the known region at step $6$, and by step $14$ the wall and door on the right have appeared, while the same region is still blurry without CNS.
Figure~\ref{fig:app-cns-revisit} compares revisit error maps of our full model and the model trained without CNS on three evaluation scenes.

\clearpage
\section{Additional Quantitative Results}
\label{sec:app-quant}
\suppressfloats[t]

\subsection{Out-of-Distribution Scenes}
\label{sec:app-quant-ood}

Table~\ref{tab:app-ood} evaluates all methods on the $50$ out-of-distribution first frames with the protocol of Section~\ref{sec:app-eval}, where motion fidelity uses $10$ of these first frames at five speeds.
Revisit PSNR and SSIM are lower than on the evaluation set for all methods, but our method remains the best on every metric, and its residual drift ($7.6$\,px) is less than half that of the strongest baseline ($16.5$\,px).

\begin{table}[t]
\caption{\textbf{Comparison on out-of-distribution scenes.} Same metrics as Table~\ref{tab:injection}, on $50$ first frames whose scenes and styles do not appear in the training game.}
\label{tab:app-ood}
\centering
\footnotesize
\setlength{\tabcolsep}{3pt}
\begin{tabular*}{\textwidth}{@{\extracolsep{\fill}}lcccccccc@{}}
\toprule
& \multicolumn{2}{c}{Motion fidelity} & Direction & \multicolumn{5}{c}{Revisit consistency} \\
\cmidrule(lr){2-3}\cmidrule(lr){4-4}\cmidrule(lr){5-9}
Method & slope & $R^2$ & angle$\downarrow$ & PSNR$\uparrow$ & SSIM$\uparrow$ & NCC$\uparrow$ & DINO$\uparrow$ & drift$\downarrow$ \\
\midrule
Wan + Prompt     & 0.831 & $-$2.39 & 4.66 & 17.64 & 0.360 & 0.664 & 0.912 & 30.3 \\
Wan + Linear     & 0.814 & 0.470   & 2.60 & 16.64 & 0.359 & 0.605 & 0.883 & 47.7 \\
Wan + AdaLN      & 0.762 & $-$0.49 & 1.71 & 18.44 & 0.376 & 0.721 & 0.921 & 16.5 \\
Wan + Cross-Attn & 0.762 & $-$2.90 & 5.41 & 17.66 & 0.363 & 0.635 & 0.902 & 36.0 \\
\midrule
Ours & \textbf{0.922} & \textbf{0.926} & \textbf{0.94} & \textbf{19.67} & \textbf{0.417} & \textbf{0.809} & \textbf{0.928} & \textbf{7.6} \\
\bottomrule
\end{tabular*}
\end{table}

\subsection{Motion Scaling per Speed}
\label{sec:app-quant-scaling}

Table~\ref{tab:app-ratio} lists the median ratio between the measured and commanded displacement at each speed, i.e., the medians in Figure~\ref{fig:app-scaling}.
Our ratios stay between $0.96$ and $0.99$.
The baselines overshoot slow commands and undershoot fast ones.
For Prompt and Cross-Attn, the ratio falls in inverse proportion to the commanded speed, since their measured displacement does not change with it.

\begin{table}[!ht]
\caption{\textbf{Measured-to-commanded displacement ratio per speed.} Median over $100$ first frames of the evaluation set. The ideal value is $1$, and the closest value in each column is in bold.}
\label{tab:app-ratio}
\centering
\footnotesize
\begin{tabular}{lccccc}
\toprule
Method & $0.5{\times}$ & $0.75{\times}$ & $1.0{\times}$ & $1.25{\times}$ & $1.5{\times}$ \\
\midrule
Wan + Prompt     & 2.05 & 1.36 & 1.02 & 0.82 & 0.68 \\
Wan + Linear     & 1.14 & \textbf{1.00} & 0.90 & 0.84 & 0.76 \\
Wan + AdaLN      & 1.16 & 1.07 & 0.95 & 0.75 & 0.60 \\
Wan + Cross-Attn & 1.80 & 1.20 & 0.90 & 0.72 & 0.60 \\
\midrule
Ours             & \textbf{0.98} & 0.99 & \textbf{0.99} & \textbf{0.99} & \textbf{0.96} \\
\bottomrule
\end{tabular}
\end{table}

\subsection{Noise Schedule Design}
\label{sec:app-schedule}

\begin{figure}[!ht]
\centering
\includegraphics[width=0.55\textwidth]{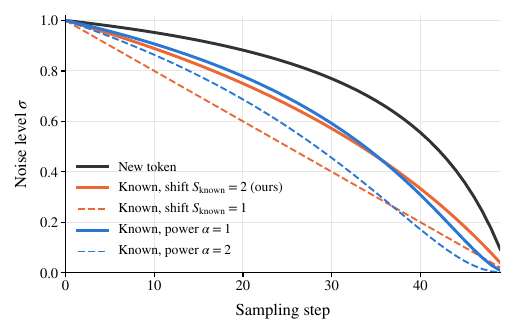}
\caption{\textbf{Noise level during sampling.} A newly revealed token follows the original schedule of Wan2.2 ($S{=}5$), while a known token ($\kappa{=}1$) follows a lower curve under each CNS variant.}
\label{fig:app-cns-schedule}
\end{figure}

Besides the shift-based schedule of Eq.~\ref{eq:cns}, we test a power-law schedule $\sigma_{kij}=\sigma^{\,1+\alpha\kappa_{kij}}$, which also keeps newly revealed tokens on the original schedule.
Figure~\ref{fig:app-cns-schedule} shows the resulting noise level of a known token for each variant, and Table~\ref{tab:app-schedule} reports the results on Wan2.2.
Every coordinate-conditioned schedule reaches a higher revisit PSNR than the uniform schedule in this comparison.
We use the shift-based form because it reuses the shifted schedule of each backbone and needs a single hyperparameter, $S_{\mathrm{known}}$.

\begin{table}[!ht]
\caption{\textbf{Noise schedule variants on Wan2.2.} Revisit consistency and direction error on the evaluation set.}
\label{tab:app-schedule}
\centering
\footnotesize
\begin{tabular}{lccc}
\toprule
Schedule & PSNR$\uparrow$ & drift$\downarrow$ & angle$\downarrow$ \\
\midrule
Uniform (without CNS)                  & 22.81 & 11.7 & \textbf{0.67} \\
Power, $\alpha{=}1$                    & 23.65 & 9.1  & 0.71 \\
Power, $\alpha{=}2$                    & 23.57 & 9.5  & 0.77 \\
Shift, $S_{\mathrm{known}}{=}1$        & 23.20 & 10.7 & 0.87 \\
Shift, $S_{\mathrm{known}}{=}2$ (ours) & \textbf{23.94} & \textbf{7.8}  & 0.68 \\
\bottomrule
\end{tabular}
\end{table}

\subsection{Training versus Sampling}
\label{sec:app-trainsample}

CNS changes both the noise applied during training and the schedule used during sampling.
To separate the two, we cross the training schedule with the sampling schedule.
Sampling a model trained without CNS with the CNS schedule leaves revisit PSNR unchanged ($22.82$ vs.\ $22.81$\,dB), and sampling the CNS-trained model with the uniform schedule keeps its revisit PSNR ($24.06$ vs.\ $23.94$\,dB).
The effect of CNS therefore comes from training rather than from the sampler alone.

\section{Implementation Details}
\label{sec:app-impl}

\subsection{ActionRoPE and CNS}
\label{sec:app-impl-arope}

Table~\ref{tab:app-impl} summarizes the settings of each backbone.
The displacement $\Delta_k$ is averaged over the video frames that form latent frame $k$, and $s$ is the number of pixels covered by one spatial token after the VAE and the patch embedding.
The world coordinates of Eq.~\ref{eq:world_coordinate} are added to the spatial RoPE positions of each backbone, and for MiniMax-H3, whose spatial positions are not spaced by one, $\Delta_k/s$ is scaled by the same spacing.
The knownness $\kappa$ is computed on the latent grid; for the token-wise timesteps, it is averaged over the latent positions of each token, so tokens on the footprint boundary receive intermediate values.
The mask channel is a zero-initialized layer that maps $\kappa$ to the hidden width and adds it to the patch embedding, i.e., a single-channel convolution with the patch-embedding kernel on Wan2.2 and SkyReels-V2, and a linear layer on LTX-2.5 and MiniMax-H3.
For CNS, $S$ is the default sampling shift of each backbone, and we keep the ratio $S_{\mathrm{known}}/S=0.4$ found on Wan2.2 for the other backbones.

\begin{table}[!ht]
\caption{\textbf{Per-backbone settings.} $s$: pixels per spatial token. Mask: parameters of the known/new mask channel. $S$, $S_{\mathrm{known}}$: CNS shifts. CFG: classifier-free guidance scale at sampling.}
\label{tab:app-impl}
\centering
\footnotesize
\setlength{\tabcolsep}{3pt}
\begin{tabular*}{\textwidth}{@{\extracolsep{\fill}}lcccccccc@{}}
\toprule
Backbone & $s$ & Mask & $S$ & $S_{\mathrm{known}}$ & Fine-tuning & LR & Batch & Steps / CFG \\
\midrule
SkyReels-V2-DF 1.3B & 16 & 7{,}680  & 5    & 2   & full          & $10^{-5}$ & 16 & 50 / 1 \\
Wan2.2-TI2V 5B      & 32 & 15{,}360 & 5    & 2   & full          & $10^{-5}$ & 16 & 50 / 1 \\
LTX-2.5 22B         & 32 & 8{,}192  & 7.77 & 3.1 & LoRA, rank 128 & $10^{-4}$ & 16 & 30 / 3 \\
MiniMax-H3 33B      & 32 & 10{,}752 & 12   & 4.8 & LoRA, rank 128 & $10^{-4}$ & 8  & 50 / 1 \\
\bottomrule
\end{tabular*}
\end{table}

\subsection{Training and Sampling}
\label{sec:app-impl-train}

All models start from the public pretrained weights and are trained for $5{,}000$ steps with AdamW, a weight decay of $0.01$, a linear warm-up followed by a constant learning rate, bf16 precision and gradient clipping at $1.0$.
The flow-matching loss excludes the first latent frame, and for our models, newly revealed tokens receive a weight of $2$.
Text prompts describe only the scene, using one of $18$ scene descriptions of the game, and the scene description is dropped with probability $0.1$.
At evaluation, all models use this scene-free prompt, the same first frames and commands, and a fixed random seed.
Sampling uses the number of steps and guidance scale in Table~\ref{tab:app-impl}, with one model evaluation per step except for the guided LTX-2.5.

\subsection{Baselines}
\label{sec:app-baselines}

All baselines are built on the same Wan2.2 backbone with standard RoPE and no mask channel, and are trained with the same data, learning rate, batch size, number of steps and seed as our model.
All of them are driven by the same per-frame camera displacement as our model (Section~\ref{sec:exp-setup}), which each baseline converts into its own action format; Prompt and Cross-Attn quantize it into discrete direction labels and therefore keep the direction but not the speed.
Since they have no known/new partition, they use a uniform loss weight.
Newly added layers are zero-initialized, so each model starts from the pretrained backbone.
\begin{itemize}
\item \textbf{Prompt}~\citep{incantation} appends the action of each latent frame to its text prompt, as one of nine labels (idle and eight directions), and the tokens of each latent frame attend only to their own prompt. It adds no parameters and receives the direction but not the speed.
\item \textbf{Linear}~\citep{reactivegwm} maps the cumulative displacement $\Delta_k/s$ through one bias-free linear layer per block and adds the result to the tokens of latent frame $k$ at the input of each of the $30$ blocks ($0.18$M parameters).
\item \textbf{AdaLN}~\citep{alayaworld} encodes the cumulative displacement and its per-frame increment with sinusoidal features, maps them with an MLP, and adds the result to the timestep modulation of every block, per latent frame ($66.5$M parameters).
\item \textbf{Cross-Attn}~\citep{matrixgame3} follows the keyboard branch of Matrix-Game~3, where each latent frame is represented by a four-key vector (up, down, left, right) and the video tokens attend to these action tokens along time in the first $15$ blocks, between the text cross-attention and the feed-forward layer ($98.6$M parameters). Like Prompt, it receives the direction but not the speed.
\end{itemize}

\section{Dataset}
\label{sec:app-data}

The recordings consist of $50.9$ hours of automated exploration in two sessions, one with the in-game day--night cycle and weather ($34.55$\,h) and one with fixed daytime and clear weather ($16.39$\,h).
The engine logs the player position at $20$\,Hz, and we map it to screen displacement with a calibrated linear map that also accounts for terrain height.
We fit the scale of this map against the background motion measured in the videos ($R^2$ of $0.99$ horizontally and $0.89$ vertically).
As a check on the action labels, the eight-way movement direction predicted from the logs agrees with that measured by optical flow on $87.7\%$ of $42{,}456$ samples, and on $98.0\%$ within one neighboring direction.

We remove corrupted segments, teleport windows, periods in which the player is stuck for at least $2$\,s, and low-light frames, leaving $33.35$ usable hours.
These are cut into $20{,}882$ clips of $81$ frames with a stride of $84$ frames and resized to $832{\times}480$.
The dataset further contains $1{,}900$ transition clips that splice two scenes.
Since adjacent clips overlap heavily, we split training and evaluation by entire exploration segments, and exclude $16$ buffer segments from both.
From the $20{,}983$ clips of the training split, we remove one clip with invalid labels, $257$ transition clips whose source clips lie in the evaluation split, and $2{,}197$ clips in which the game world runs slower than $0.8{\times}$ real time, which leaves $18{,}528$ training clips.
The evaluation set contains the $982$ held-out clips with valid labels that run at least at $0.9{\times}$ real time.

\section{Evaluation Protocol}
\label{sec:app-eval}

\paragraph{Commands.}
The nominal speed ($1.0{\times}$) is the median walking speed in the training data, $4.33$\,px per frame horizontally.
Motion fidelity uses the first $100$ evaluation first frames and commands a straight walk to the right for $80$ frames at $0.5$, $0.75$, $1.0$, $1.25$ and $1.5{\times}$ this speed.
Directional compliance commands the same walk at $1.0{\times}$ on all $982$ first frames.
Revisit consistency commands a walk to the right for $40$ frames followed by a walk to the left for $40$ frames, so that the commanded camera position at the final frame equals the initial one.

\paragraph{Measuring motion.}
We detect SIFT features on contrast-enhanced grayscale frames, excluding a $160{\times}160$-pixel box at the screen center where the player stays, and fit an affine motion with RANSAC.
The displacement between the first and the last frame is measured directly, and through intermediate frames when direct matching fails.
Clips whose motion cannot be measured, mostly very dark first frames, are excluded from the medians.

\paragraph{Metrics.}
For motion fidelity, we regress the measured displacement on the commanded one through the origin and report the slope.
$R^2$ is computed around the mean of the measured displacement, so it becomes negative when the measured displacement does not vary with the command.
For directional compliance, we report the median absolute angle between the measured and commanded displacement.
For revisit consistency, we compare the final frame with the first frame outside the central box.
Before computing PSNR, we match the mean and standard deviation of each color channel of the final frame to the first frame, so that global brightness changes are not counted.
SSIM uses an $11{\times}11$ Gaussian window with $\sigma{=}1.5$ on grayscale frames, and NCC is computed on grayscale frames.
DINO is the cosine similarity between the DINOv2-base class tokens of the two frames, with the central box filled with gray.
The residual drift is the magnitude of the measured displacement between the first and the final frame.

\paragraph{Out-of-distribution first frames.}
The $50$ out-of-distribution first frames depict scenes and art styles that do not appear in the game, such as cyberpunk cities, deserts, oil paintings and pixel art, each with the player at the screen center.
They are evaluated with the same commands and metrics, with motion fidelity on the first $10$ of them.

\section{Limitations}
\label{sec:app-limit}

\ourmethod{} assumes that the static world translates approximately rigidly in the image plane, which holds for top-down and isometric cameras that follow the player.
It does not model zoom or strong parallax, and moving objects do not follow the world coordinates.
The knownness $\kappa$ covers only the first-frame footprint, so regions that become visible later are not treated as known when the camera returns to them.

\end{document}